\pdfoutput=1

\documentclass[11pt]{article}

\usepackage{acl}

\usepackage{times}
\usepackage{latexsym}

\usepackage[T1]{fontenc}

\usepackage[utf8]{inputenc}

\usepackage{microtype}

\usepackage{amsmath,amsfonts,bm}

\def\eqref#1{equation~\ref{#1}}

\def\1{\bm{1}}

\DeclareMathAlphabet{\mathsfit}{\encodingdefault}{\sfdefault}{m}{sl}
\SetMathAlphabet{\mathsfit}{bold}{\encodingdefault}{\sfdefault}{bx}{n}

\DeclareMathOperator*{\argmax}{arg\,max}

\usepackage{hyperref}
\usepackage{url}

\usepackage{pifont}
\usepackage{url}
\usepackage[utf8]{inputenc} 
\usepackage[T1]{fontenc}    
\usepackage{url}            
\usepackage{booktabs}       
\usepackage{amsfonts}       
\usepackage{nicefrac}       
\usepackage{microtype}      
\usepackage{times}
\usepackage{xcolor}
\usepackage{color}
\usepackage{soul}
\usepackage[utf8]{inputenc}
\usepackage{graphicx} 
\usepackage{wrapfig}
\usepackage{amssymb}
\usepackage{amsmath}
\usepackage{multirow}
\usepackage{float}
\usepackage{subcaption}
\usepackage{arydshln}

\usepackage{times}
\usepackage{epsfig}
\usepackage{graphicx}
\usepackage{amsmath}
\usepackage{amssymb}
\usepackage{textcomp}
\usepackage{wrapfig}

\title{Multimodal Dialogue State Tracking}

\author{Hung Le${^\dag}{^\ddag}{^\S}$, Nancy F. Chen$^\S$, Steven C.H. Hoi${^\dag}{^\ddag}$ \\
  $^{\dag}$Salesforce Research Asia \\
  $^{\ddag}$Singapore Management University\\
  $^\S$Agency for Science, Technology and Research (A*STAR) \\
  \texttt{\{hungle,shoi\}@salesforce.com, nfychen@i2r.a-star.edu.sg}
  }

\begin{document}
\maketitle
\begin{abstract}
Designed for tracking user goals in dialogues, a dialogue state tracker is an essential component in a dialogue system. However, the research of dialogue state tracking has largely been limited to unimodality, in which slots and slot values are limited by knowledge domains (e.g. restaurant domain with slots of restaurant name and price range) and are defined by specific database schema. In this paper, we propose to extend the definition of dialogue state tracking to multimodality. Specifically, we introduce a novel dialogue state tracking task to track the information of visual objects that are mentioned in video-grounded dialogues. Each new dialogue utterance may introduce a new video segment, new visual objects, or new object attributes and a state tracker is required to update these information slots accordingly. 
We created a new synthetic benchmark and designed a novel baseline, Video-Dialogue Transformer Network (VDTN), for this task. VDTN combines both object-level features and segment-level features and learns contextual dependencies between videos and dialogues to generate multimodal dialogue states. We optimized VDTN for a state generation task as well as a self-supervised video understanding task which recovers video segment or object representations. Finally, we trained VDTN to use the decoded states in a response prediction task. Together with comprehensive ablation and qualitative analysis, we discovered interesting insights towards building more capable multimodal dialogue systems. 
\end{abstract}

\section{Introduction}
The main goal of dialogue research 
is to develop intelligent agents that can assist humans through conversations. 
For example, 
a dialogue agent can be tasked to help users to find a restaurant based on their preferences of price ranges and food choices. 
A crucial part of a dialogue system is Dialogue State Tracking (DST), which is responsible for tracking and updating user goals in the form of dialogue states, including a set of \emph{(slot, value)} pairs such as \emph{(price, ``moderate'')} and \emph{(food, ``japanese'')}.
Numerous machine learning approaches have been proposed to tackle DST, including fixed-vocabulary models \citep{ramadan2018large, lee2019sumbt} and open-vocabulary models \citep{lei2018sequicity, wu-etal-2019-transferable, Le2020Non-Autoregressive}, for either single-domain \citep{wen-etal-2017-network} or multi-domain dialogues \citep{eric-etal-2017-key, budzianowski-etal-2018-multiwoz}. 

\begin{figure*}[h]
	\centering
	\resizebox{1.0\textwidth}{!} {
	\includegraphics{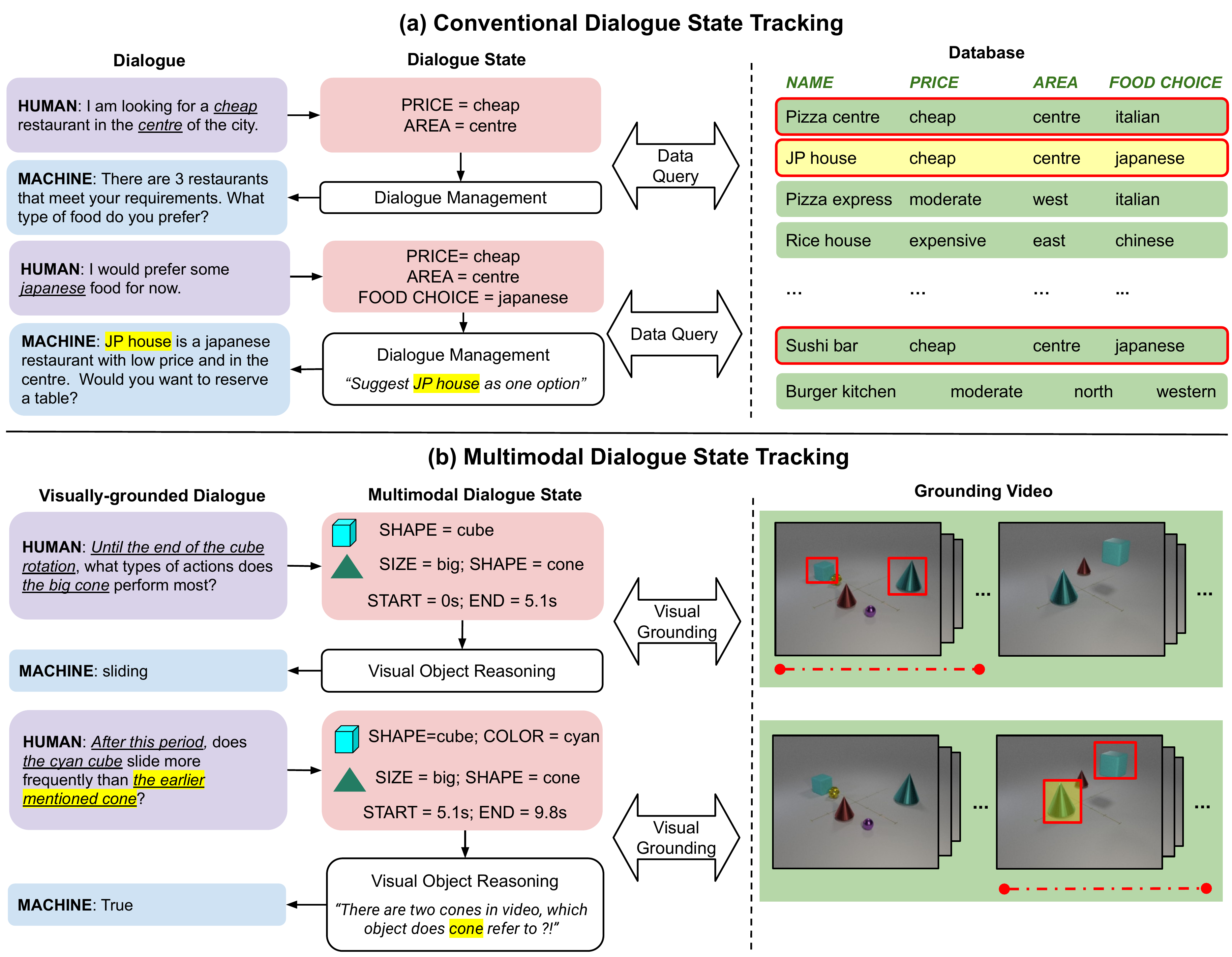}
	}
	\caption{
	\textbf{Multimodal Dialogue State Tracking (MM-DST):}
	We proposed to extend the traditional DST from unimodality to multimodality. 
	Compared to traditional DST (a), MM-DST (b) define dialogue states, consisting of slots and slots values for visual objects that are mentioned in dialogues. 
	}
	\label{fig:mm_dst}
\end{figure*}

However, the research of DST has largely limited the scope of dialogue agents to unimodality. 
In this setting, the slots and slot values are defined by the knowledge domains (e.g. restaurant domain) and database schema (e.g. data tables for restaurant entities). 
The ultimate goal of dialogue research towards building artificial intelligent assistants necessitates DST going beyond unimodal systems. 
In this paper, we propose Multimodal Dialogue State Tracking (MM-DST) that extends the DST task in a multimodal world.
Specifically, MM-DST extends the scope of dialogue states by defining slots and slot values for \emph{visual objects} that are mentioned in visually-grounded dialogues. 
For research purposes, following \citep{alamri2019audiovisual}, we limited visually-grounded dialogues as ones with a grounding video input and the dialogues contain multiple turns of (question, answer) pairs about this video. 
Each new utterance in such dialogues may focus on a new video segment, new visual objects, or new object attributes, and the tracker is required to update the dialogue state accordingly at each turn. 
An example of MM-DST can be seen in Figure \ref{fig:mm_dst}.


Toward MM-DST, we developed a synthetic benchmark based on the CATER universe \citep{Girdhar2020CATER}. We also introduced Video-Dialogue Transformer Network (VDTN), a neural network architecture that combines both object-level features and segment-level features in video and learns contextual dependencies between videos and dialogues. 
Specifically, we maintained the information granularity of visual objects, embedded by object classes and their bounding boxes and injected with segment-level visual context. 
VDTN enables interactions between each visual object representation and word-level representation in dialogues to decode dialogue states. 
To decode multimodal dialogue states, we adopted a decoding strategy inspired by the Markov decision process in traditional DST \citep{YOUNG2010150}.
In this strategy, a model learns to decode the state at a dialogue turn based on the predicted/ observed dialogue state available from the last dialogue turn. 

Compared to the conventional DST, MM-DST involves the new modality from visual inputs. 
Our experiments show that simply combining visual and language representations in traditional DST models results in poor performance.
Towards this challenge, we enhanced VDTN with self-supervised video understanding tasks which recovers object-based or segment-based representations. 
Benchmarked against strong unimodal DST models, we observed significant performance gains from VDTN.
We provided comprehensive ablation analysis to study the efficacy of VDTN models.  
Interestingly, we also showed that using decoded states brought performance gains in a dialogue response prediction task, supporting our motivation for introducing multimodality into DST research. 

\section{Multimodal Dialogue State Tracking}
\label{sec:mm_dst_task}
\paragraph{Traditional DST.}
As defined by \cite{mrksic2017neuralbelief}, the traditional DST includes an input of dialogue $\mathcal{D}$ and a set of slots $\mathcal{S}$ to be tracked from turn to turn. 
At each dialogue turn $t$, we denote the dialogue context as $\mathcal{D}_t$, containing all utterances up to the current turn. 
The objective of DST is for each turn $t$, predict a value $v^t_i$ of each slot $s_i$ from a predefined set $\mathcal{S}$, conditioned by the dialogue context $\mathcal{D}_t$.  
We denote the dialogue state at turn $t$ as $\mathcal{B}_t = \{(s_i, v^t_i)\} |_{i=1}^{i=|\mathcal{S}|}$.
Note that a majority of traditional DST models assume slots are conditionally independent, given the dialogue context \citep{zhong-etal-2018-global, budzianowski-etal-2018-multiwoz, wu-etal-2019-transferable, lee2019sumbt, gao2019dialog}.
The learning objective is defined as: 
\begin{align}
    \hat{\mathcal{B}}_t &= \argmax_{\mathcal{B}_t} P(\mathcal{B}_t | \mathcal{D}_t, \theta) \nonumber \\
    &= \argmax_{\mathcal{B}_t} \prod_{i}^{|\mathcal{S}|} P(v_i^t | s_i, \mathcal{D}_t,\theta)
     \label{eq:dst_objective}
\end{align}
\paragraph{Motivation to Multimodality.}
Yet, the above definition of DST are still limited to unimodality and our ultimate goal of building intelligent dialogue agents, ideally with similar level of intelligence as humans, inspires us to explore mulitmodality. 
In neuroscience literature, several studies have analyzed how humans can perceive the world in visual context. \cite{bar2004visual, xu2009selecting} found that humans can recognize multiple visual objects and how their contexts, often embedded with other related objects, facilitate this capacity. 

Our work is more related to the recent study \citep{fischer2020context} which focuses on human capacity to create temporal stability across multiple objects. 
The multimodal DST task is designed to develop multimodal dialogue systems that are capable of maintaining discriminative representations of visual objects over a period of time, segmented by dialogue turns. 
While computer science literature has focused on related human capacities in intelligent systems, they are mostly limited to vision-only tasks e.g. \citep{he2016deep, ren2015faster} or QA tasks e.g. \citep{antol2015vqa, jang2017tgif} but not in a dialogue task. 

Most related work in the dialogue domain is \citep{pang2020visual} and almost concurrent to our work is \citep{kottur-etal-2021-simmc}. However, \citep{kottur-etal-2021-simmc} is limited to a single object per dialogue, and \citep{pang2020visual} extends to multiple objects but does not require to maintain an information state with component slots for each object. Our work aims to complement these directions and address their limitations with a novel definition of multimodal dialogue state. 

\paragraph{Multimodal DST (MM-DST).}
To this end, we proposed to extend conventional dialogue states.
First, we use visual object identities themselves as a component of the dialogue state to enable the perception of multiple objects.
A dialogue state might have one or more objects and a dialogue system needs to update the object set as the dialogue carries on. 
Secondly, for each object, we define slots that represent the information state of objects in dialogues (as denoted by \cite{fischer2020context} as ``content'' features of objects memorized by humans). 
The value of each slot is subject-specific and updated based on the dialogue context of the corresponding object. 
This definition of DST is closely based on the above well-studied human capacities while complementing the conventional dialogue research \citep{YOUNG2010150, mrksic2017neuralbelief}, and more lately multimodal dialogue research \citep{pang2020visual, kottur-etal-2021-simmc}. 

We denote a grounding visual input in the form of a video $\mathcal{V}$ with one or more visual objects $o_j$.
We assume these objects are semantically different enough (by appearance, by characters, etc.) such that each object can be uniquely identified (e.g. by an object detection module $\omega$). 
The objective of MM-DST is for each dialogue turn $t$, predict a value $v_i^t$ of each slot $s_i \in \mathcal{S}$ for each object $o_j \in \mathcal{O}$.
We denote the dialogue state at turn $t$ as 
$\mathcal{B}_t = |\{(o_j, s_i, v_{i,j}^t)\}|_{i=1, j=1}^{i=|\mathcal{S}|, j=|\mathcal{O}|}$.
 Assuming all slots are conditionally independent given dialogue and video context, the learning objective is extended from Eq. (\ref{eq:dst_objective}):
 
{ \small
\begin{align}
    \hat{\mathcal{B}}_t &= \argmax_{\mathcal{B}_t} P(\mathcal{B}_t | \mathcal{D}_t, \mathcal{V}, \theta) \nonumber \\
   &= \argmax_{\mathcal{B}_t} \prod_{j}^{|\mathcal{O}|} \prod_{i}^{|\mathcal{S}|} P(v_{i,j}^t | s_i, o_j, \mathcal{D}_t, \mathcal{V}, \theta) P(o_j | \mathcal{V}, \omega) \nonumber
\end{align}
 }
One limitation of the current representation is the absence of temporal placement of objects in time. 
Naturally humans are able to associate objects and their temporal occurrence over a certain period. 
Therefore, we defined two temporal-based slots: $s_\mathrm{start}$ and $s_\mathrm{end}$, denoting the start time and end time of the video segment that an object can be located by each dialogue turn.  
In this work, we assume that a dialogue turn is limited to a single continuous time span, and hence, $s_\mathrm{start}$ and $s_\mathrm{end}$ can be defined turn-wise, identically for all objects.  
While this is a strong assumption, we believe it covers a large portion of natural conversational interactions.
An example of multimodal dialogue state can be seen in Figure \ref{fig:mm_dst}.

\begin{figure*}[t]
	\centering
	\resizebox{1.0\textwidth}{!} {
	\includegraphics{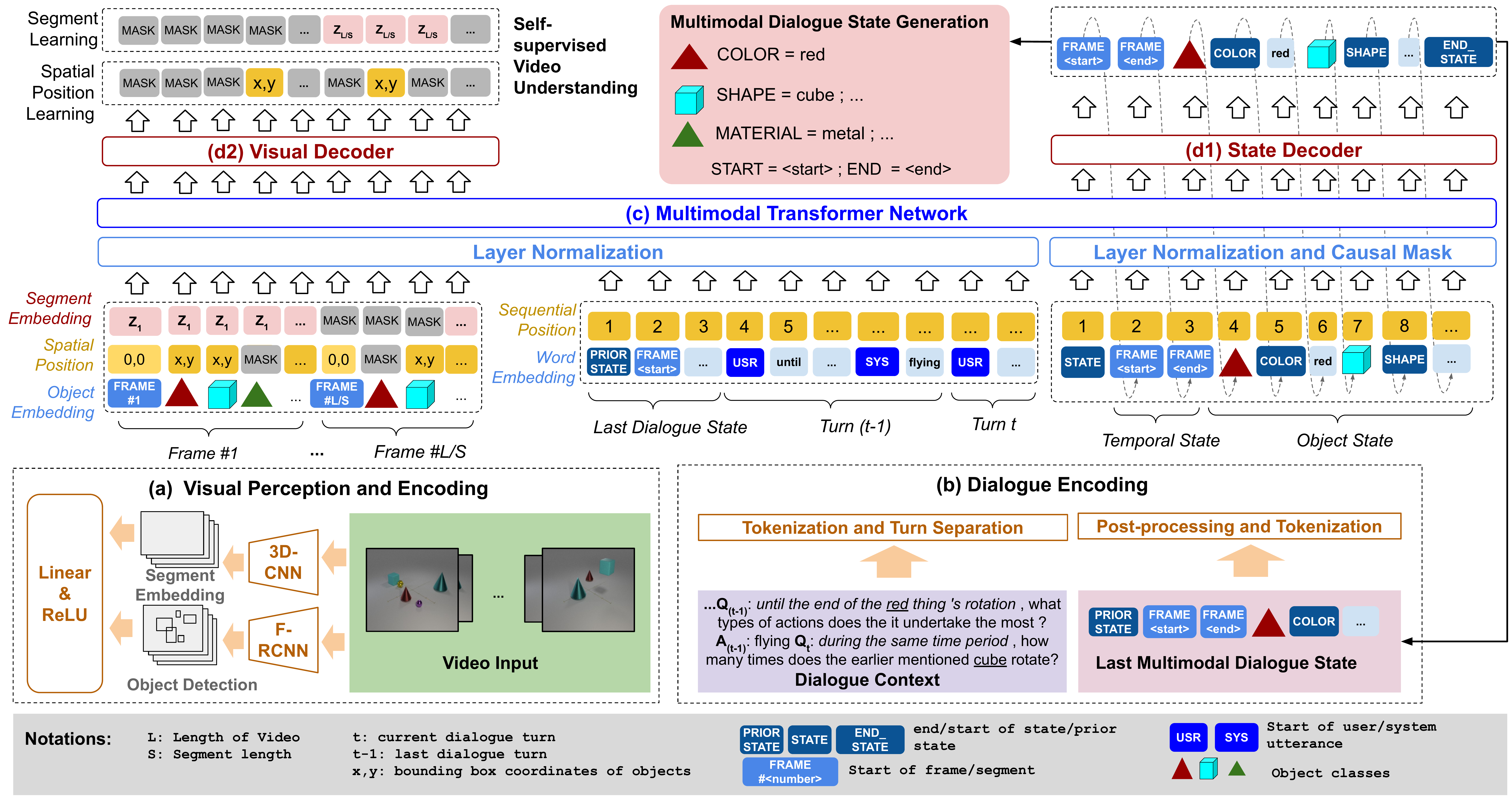}
	}
	\caption{
	\textbf{Video-Dialogue Transformer Network(VDTN)} has 4 key components:
	(a) \emph{Visual Perception and Encoder}
	(Section \ref{subsec:video_encode})
	(b) \emph{Dialogue Encoder} 
	(Section \ref{subsec:dial_encode})
	(c) \emph{Transformer Network} 
	(Section \ref{subsec:mm_transformer})
	(d1) \emph{State Decoder} 
	(Section \ref{subsec:state_video_decode})
	and
	(d2) \emph{Visual Decoder}
	(Section \ref{subsec:state_video_decode})
	}
	\label{fig:model}
\end{figure*}

\section{Video-Dialogue Transformer Network}
\label{sec:method}
A naive adaptation of conventional DST to MM-DST is to directly combine visual features extracted by a pretrained 3D-CNN model. 
However, as shown in our experiments, this extension of conventional DST results in poor performance and does not address the challenge of MM-DST.
In this paper, we established a new baseline, denoted as Video-Dialogue Transformer Network (VDTN) (Refer to Fig. \ref{fig:model} for an overview):

\subsection{Visual Perception and Encoder}
\label{subsec:video_encode}
\paragraph{Visual Perception.} This module encodes videos at both frame-level and segment-level representations. 
Specifically, we used a pretrained Faster R-CNN model \citep{ren2015faster} 
to extract object representations. We used this model to output the bounding boxes and object identifiers (object classes) in each video frame of the video.
For an object $o_j$, we denoted the four values of its bounding boxes as ${bb}_j = (x^1_j, y^1_j, x^2_j, y^2_j)$ and $o_j$ as the object class itself. We standardized the video features by extracting features of up to $N_{obj}=10$ objects per frame and normalizing all bounding box coordinates by the frame size. Secondly, we used a pretrained ResNeXt model \citep{xie2017aggregated} to extract the segment-level representations of videos, denoted as $z_m \in \mathbb{R}^{2048}$ for a segment $m$. 
Practically, we followed the best practice in computer vision by using a temporal sliding window with strides to sample video segments and passed segments to ResNeXt model to extract features. To standardize visual features, we use the same striding configuration $N_{stride}$ to sub-sample segments for ResNeXt and frames for Faster R-CNN models. 

\paragraph{Visual Representation.} Note that we do not finetune the visual feature extractors in VDTN and keep the extracted features fixed. 
To transform these features into VDTN embedding space, we first concatenated all object identity tokens \emph{OBJ<class>} of all frames.
An object identity token \emph{OBJ<class>} is the code name of the object class (e.g. a class of small blue metal cones) that a visual object can be unique identified (See Figure \ref{fig:model}).
Frames are separated by a special token \emph{FRAME<number>}, where \emph{<number>} is the temporal order of the frame.
This results in a sequence of tokens $X_{obj}$ of length $L_{obj} = (N_{obj}+1) \times (|\mathcal{V}|/N_{stride})$ where $|\mathcal{V}|$ is the number of video frames. 
Correspondingly, we concatenated bounding boxes of all objects, and used a zero vector in positions of \emph{FRAME<number>} tokens.
We denoted this sequence as $X_{bb} \in \mathbb{R}^{L_{obj} \times 4}$ where the dimension of $4$ is for the bounding box coordinates $(x^1, y^1, x^2, y^2)$.
Similarly, we stacked each ResNeXt feature vector by $(N_{obj}+1)$ for each segment, and obtained a sequence $X_{cnn} \in \mathbb{R}^{L_{obj} \times {2048}}$.

\paragraph{Visual Encoding.}
We passed each of $X_{bb}$ and $X_{cnn}$ to a linear layer with ReLU activation to map their feature dimension to a uniform dimension $d$. 
We used a learnable embedding matrix to embed each object identity in $X_{obj}$, resulting in embedding features of dimensions $d$. 
The final video input representation is the summation of above vectors, denoted as $Z_{V}=Z_{obj}+Z_{bb}+Z_{cnn} \in \mathbb{R}^{L_{obj} \times d}$.

\subsection{Dialogue and State Encoder}
\label{subsec:dial_encode}
\paragraph{Dialogue Encoding.}
Another encoder encodes dialogue into continuous representations.
Given a dialogue context $\mathcal{D}_t$, we tokenized all dialogue utterances into sequences of words, separated by special tokens \emph{USR} for human utterance and \emph{SYS} for system utterance. 
We used a trainable embedding matrix and sinusoidal positional embeddings to embed this sequence into $d$-dimensional vectors. 

\paragraph{Flattening State into Sequence.}
Similar to the recent work in traditional DST \citep{lei2018sequicity, le-etal-2020-uniconv, zhang-etal-2020-find}, we are motivated by the DST decoding strategy following a Markov principle and used the dialogue state of the last dialogue turn $\mathcal{B}_{t-1}$ as an input to generate the current state $\mathcal{B}_t$.
Using the same notations from Section \ref{sec:mm_dst_task}, we can represent $B_t$ into a sequence of $o_j, s_i$, and $v^t_{i,j}$ tokens, such as ``\emph{OBJ4 shape cube OBJ24 size small color red}''. 
This sequence is then concatenated with utterances from $\mathcal{D}_t$, separated by a special token \emph{PRIOR\_STATE}.
We denoted the resulting sequence as $X_{ctx}$ which is passed to the embedding matrix and positional encoding as described above.
As we showed in our experiments, to encode dialogue context, this strategy needs only a few dialogue utterances (that are closer to the current turn $t$) and $\mathcal{B}_{t-1}$, rather than the full dialogue history from turn $1$. 
Therefore, dialogue representations $Z_{ctx}$ have more compressed dimensions of $|X_{ctx}| \times d$ where $|X_{ctx}| < |\mathcal{D}_t|$.

\subsection{Multimodal Transformer Network}
\label{subsec:mm_transformer}
We concatenated both video and dialogue representations, denoted as $Z_{VD} = [Z_V ; Z_D]$. $Z_{VD}$ has a length of $L_{obj}+L_{ctx}$ and embedding dimension $d$. 
We pased $Z_{VD}$ to a  vanilla Transformer network \citep{vaswani17attention} through multiple multi-head attention layers with normalization \citep{ba2016layer} and residual connections \citep{he2016deep}. 
Each layer allows multimodal interactions between object-level representations from videos and word-level representations from dialogues.

\subsection{Dialogue State Decoder and Self-supervised Video Denoising Decoder}
\label{subsec:state_video_decode}
\paragraph{State Decoding.}
This module decodes dialogue state sequence auto-regressively, i.e. each token is conditioned on all dialogue and video representations as well as all tokens previously decoded.
At the first decoding position, a special token \emph{STATE} is embedded into dimension $d$ (by a learned embedding layer and sinusoidal positional encoding) and concatenated to $Z_{VD}$.
The resulting sequence is passed to the Transformer network and the output representations of \emph{STATE} are passed to a linear network layer that transforms representations to state vocabulary embedding space.
The decoder applies the same procedure for the subsequent positions to decode dialogue states auto-regressively. 
Denoting $b_{k,t}$ as the $k^{th}$ token in $\mathcal{B}_t$, i.e. token of slot, object identity, or slot value, we defined the DST loss function as the negative log-likelihood:
\begin{align*}
    \mathcal{L}_{dst} = - \sum \log P(b_{k,t} | b_{<k,t}, X_{ctx}, X_{obj})
\end{align*}
Note that this decoder design partially avoids the assumption of conditionally independent slots.
During test time, we applied beam search to decode states with the maximum length of 25 tokens in all models and a beam size 5. An \emph{END\_STATE} token is used to mark the end of each sequence. 
\paragraph{Visual Denoising Decoding.}
Finally, moving away from conventional unimodal DST, we proposed to enhance our DST model with a \emph{Visual Decoder} that learns to recover visual representations in a self-supervised learning task to improve video representation learning. 
Specifically, during training time, we randomly sampled visual representations and masked each of them with a zero vector. 
At the object level, in the $m^{th}$ video frame, we randomly masked a row from $X_{bb}(m) \in \mathbb{R}^{N_{obj}\times 4}$.
Since each row represents an object, we selected a row to mask by a random object index $j \in [1, N_{obj}]$ such that the same object has not been masked in the preceding frame or following frame. 
We denote the Transformer output representations from video inputs as $Z'_V \in \mathbb{R}^{L_{obj} \times d}$.
This vector is passed to a linear mapping $f_{bb}$ to bounding box features $\mathbb{R}^4$. We defined the learning objective as: 
\begin{align*}
    \mathcal{L}_{obj} &= \sum_o \displaystyle \1_\mathrm{masked} \times l(f_{bb}(Z'_{V,o}), X_{bb,o}) 
\end{align*}
where $l$ is a loss function and $ \1_\mathrm{masked}=\{0,1\}$ is a masking indicator. We experimented with both L1 and L2 loss and reported the results. 
Similarly, at the segment level, we randomly selected a segment to mask such that the preceding or following segments have not been chosen for masking:
\begin{align*}
    \mathcal{L}_{seg} &= \sum_s \displaystyle \1_\mathrm{masked} \times l(f_{cnn}(Z'_{V,s}), X_{cnn,s})
\end{align*}

\section{Experiments}
\label{sec:exps}
\subsection{Experimental Setup}

\paragraph{Dataset.} 
In existing benchmarks of multimodal dialogues such as VisDial \citep{das2017visual} and AVSD \citep{alamri2019audiovisual}, we observed that a large number of data samples contain strong distribution bias in dialogue context, in which dialogue agents can simply ignore the whole dialogue and rely on image-only features \citep{Kim_Tan_Bansal_2020}.
Another observed bias is the annotator bias that makes a causal link between dialogue context and output response actually harmful \citep{qi2020two} (as annotator's preferences are treated as a confounding factor). The above biases would obviate the need for a DST task.

To address the above biases, \citet{le-etal-2021-dvd} developed a \underline{D}iagnostic Benchmark for \underline{V}ideo-grounded \underline{D}ialogues (``DVD''), by synthetically creating dialogues that are grounded on videos from CATER videos \citep{shamsian2020learning}.
The videos contain visually simple yet highly varied objects.
The dialogues are synthetically designed with both short-term and long-term object references. These specifications remove the annotation bias in terms of object appearances in visual context and cross-turn dependencies in dialogue context. 


\paragraph{Extension from DVD \citep{le-etal-2021-dvd}}.
We generated new dialogues following \citet{le-etal-2021-dvd}'s procedures but based on an extended CATER video split \citep{shamsian2020learning} rather than the original CATER video data \citep{Girdhar2020CATER}. 
We chose the extended CATER split \citep{shamsian2020learning} as it includes additional annotations of ground-truth bounding box boundaries of visual objects in video frames. 
This annotation facilitates experiments with Faster-RCNN finetuned on CATER objects and experiments with models of perfect visual perception, i.e. $P(o_j | \mathcal{V}, \omega) \approx 1$.
As shown in \citep{le-etal-2021-dvd}, objects can be uniquely referred in utterances based on their appearance by one or more following aspects: ``size'', ``color'', ``material'', and ``shape''. 
We directly reuse these and define them as slots in our dialogue states, in addition to 2 temporal slots for $s_\mathrm{start}$ and $s_\mathrm{end}$.
We denote the new benchmark as DVD-DST and summarize the dataset in Table \ref{tab:data_main} (for more detail, please refer to Appendix \ref{app:data}).

\begin{table}[t]
\centering
\small
\resizebox{1.0\columnwidth}{!} {
\begin{tabular}{llll}
\hline
\multicolumn{1}{c}{\textbf{Split}} & \multicolumn{1}{c}{\textbf{\# Videos}} & \multicolumn{1}{c}{\textbf{\# Dialogues}} & \multicolumn{1}{c}{\textbf{\# Turns}} \\
\hline
DVD-DST-Train                              & 9300                                   & 9295                                      & 92950                                      \\
DVD-DST-Val                         & 3327                                   & 3326                                      & 33260                                  \\
DVD-DST-Test                               & 1371                                   & 1371                                      & 13710               \\
DVD-DST-All                               & 13998                                   & 13992                                      & 139920               \\
\hline
\end{tabular}
}
\caption{Summary of the DVD-DST dataset}
\label{tab:data_main}
\end{table}

\paragraph{Baselines.} 
To benchmark VDTN, we compared the model with following baseline models, including both rule-based models and trainable models:  
\begin{itemize}
    \item \emph{Q-retrieval (tf-idf)}, for each test sample, directly retrieves the training sample with the most similar question utterance and use its state as the predicted state; 
    \item \emph{State prior} selects the most common tuple of (object, slot, value) in training split and uses it as predicted states;
    \item \emph{Object (random)}, for each test sample, randomly selects one object predicted by the visual perception model and a random (slot, value) tuple (with slots and values inferred from object classes) as the predicted state; 
    \item \emph{Object (all)} is similar to the prior baseline but selects all possible objects and all possible (slot, value) tuples as the predicted state; 
    \item \emph{RNN(+Att)} uses RNN as encoder and an MLP network as decoder. Another variant of the model is enhanced with a vanilla dot-product attention at each decoding step;
    \item  We adapted and experimented with strong unimodal DST baselines, including: \emph{TRADE} \citep{wu-etal-2019-transferable}, \emph{UniConv} \citep{le-etal-2020-uniconv} and \emph{NADST} \citep{Le2020Non-Autoregressive}.
\end{itemize}

We implemented these baselines and tested them on dialogues with or without videos. 
When video inputs are applied, we embedded both object and segment-level features (See Section \ref{subsec:video_encode}). 
The video context features are integrated into baselines in the same techniques in which the original models treat dialogue context features.

\paragraph{Training.}
We trained VDTN by jointly minimizing $\mathcal{L}_{dst}$ and $\mathcal{L}_{bb/seg}$. 
We trained all models using the Adam optimizer \citep{kingma2014adam} with a warm-up learning rate period of 1 epoch and the learning rate decays up to 160 epochs. 
Models are selected based on the average $\mathcal{L}_{dst}$ on the validation set. 
To standardize model sizes, we selected embedding dimension $d=128$ for all models, and experimented with both shallow ($N=1$) and deep networks ($N=3$) (by stacking attention or RNN blocks), and $8$ attention heads in Transformer backbones. 
We implemented models in Pytorch and released the code and model checkpoints \footnote{\url{https://github.com/henryhungle/mm_dst}}. 
Refer to Appendix \ref{app:training} for more training details. 

\paragraph{Evaluation.}
We followed the unimodal DST task \citep{budzianowski-etal-2018-multiwoz, henderson2014second} and used the state accuracy metric. 
The prediction is counted as correct only when all the component values exactly match the oracle values. 
In multimodal states, there are both discrete slots (object attributes) as well as continuous slots (temporal start and end time). 
For continuous slots, we followed \citep{hu2016natural, gao2017tall} by using Intersection-over-Union (IoU) between predicted temporal segment and ground-truth segment. 
The predicted segment is counted as correct if its $IoU$ with the oracle is more than $p$, where we chose $p=\{0.5, 0.7\}$.
We reported the joint state accuracy of discrete slots only (``Joint Acc'') as well as all slot values (``Joint Acc IoU@$p$'').
We also reported the performance of component state predictions, including predictions of object identities $o_j$, object slot tuples $(o_j, s_i, v_{i,j})$, and object state tuples $(o_j, s_i, v_{i,j}) \forall s_i \in \mathcal{S}$.
Since a model may simply output all possible object identities and slot values and achieve $100$\% component accuracies, we reported the F1 metric for each of these component predictions.

\begin{table}[t]
\centering
\resizebox{1.0\columnwidth}{!} {
\begin{tabular}{lccrrrrrr}
\hline
\multicolumn{1}{c}{Model}     & Dial.   & Vid. & \multicolumn{1}{c}{\begin{tabular}[c]{@{}c@{}}Obj \\ Ident.\\ F1\end{tabular}} & \multicolumn{1}{c}{\begin{tabular}[c]{@{}c@{}}Obj\\ Slot \\ F1\end{tabular}} & \multicolumn{1}{c}{\begin{tabular}[c]{@{}c@{}}Obj \\ State \\ F1\end{tabular}} & \multicolumn{1}{c}{\begin{tabular}[c]{@{}c@{}}Joint \\ Acc \\ \end{tabular}} & \multicolumn{1}{c}{\begin{tabular}[c]{@{}c@{}}Acc \\ IoU \\ @.5\end{tabular}} & \multicolumn{1}{c}{\begin{tabular}[c]{@{}c@{}}Acc \\ IoU \\ @.7\end{tabular}} \\
\hline
Q-retrieval         & \checkmark &  -     & 6.7                                                                            & 3.3                                                                        & 2.7                                                                          & 1.0                                                                                 & 0.8                                                                                 & 0.7                                                                                 \\
State prior                  &  -      &  -     & 14.9                                                                           & 7.7                                                                        & 0.1                                                                          & 0.0                                                                                 & 0.0                                                                                 & 0.0                                                                                 \\
Object (rand.)               &   -     &\checkmark   & 19.8                                                                           & 14.1                                                                       & 0.4                                                                          & 0.0                                                                                 & 0.0                                                                                 & 0.0                                                                                 \\
Object (all)                  &  -      & \checkmark  & 60.5                                                                           & 27.2                                                                       & 1.5                                                                          & 0.0                                                                                 & 0.0                                                                                 & 0.0                                                                                 \\ \hline
RNN(V)                    &  -      & \checkmark     & 21.2                                                                           & 10.8                                                                       & 8.3                                                                          & 1.0                                                                                 & 0.1                                                                                 & 0.1                                                                                 \\
RNN(D)        & \checkmark      & -       & 57.8                                                                           & 43.3                                                                       & 38.0                                                                         & 4.8                                                                                 & 1.1                                                                                 & 0.6                                                                                 \\
RNN(V+D)        & \checkmark      & \checkmark     & 63.2                                                                           & 48.5                                                                       & 42.8                                                                         & 6.8                                                                                 & 2.6                                                                                 & 2.3                                                                                 \\
RNN(V+D)+Att & \checkmark     & \checkmark     & 73.4                                                                           & 59.0                                                                       & 46.8                                                                         & 8.5                                                                                 & 3.3                                                                                 & 2.0                                                                                 \\ \hline
TRADE (N=1)                   & \checkmark      & -      & 75.3                                                                           & 63.2                                                                       & 47.8                                                                         & 8.7                                                                                 & 2.2                                                                                 & 1.1                                                                                 \\
TRADE (N=1)                   & \checkmark      & \checkmark     & 75.8                                                                           & 63.8                                                                       & 48.0                                                                         & 9.2                                                                                 & 3.3                                                                                 & 2.5                                                                                 \\
TRADE (N=3)                   & \checkmark      & -      & 74.2                                                                           & 62.6                                                                       & 47.2                                                                         & 8.3                                                                                 & 2.1                                                                                 & 1.1                                                                                 \\
TRADE (N=3)                   & \checkmark      & \checkmark     & 76.1                                                                           & 64.5                                                                       & 48.2                                                                         & 8.9                                                                                 & 3.2                                                                                 & 2.4                                                                                 \\ \hline
UniConv (N=1)                 & \checkmark      & -      & 70.6                                                                           & 58.0                                                                       & 44.7                                                                         & 11.1                                                                                & 4.5                                                                                 & 3.2                                                                                 \\
UniConv (N=1)                 & \checkmark      & \checkmark     & 73.6                                                                           & 60.5                                                                       & 46.2                                                                         & 11.6                                                                                & 6.1                                                                                 & 5.4                                                                                 \\
UniConv (N=3)                 & \checkmark      & -      & 76.4                                                                           & 62.7                                                                       & 52.5                                                                         & 15.0                                                                                & 6.4                                                                                 & 4.6                                                                                 \\
UniConv (N=3)                 & \checkmark      & \checkmark     & 76.4                                                                           & 62.7                                                                       & 50.5                                                                         & 14.5                                                                                & 7.8                                                                                 & 7.0                                                                                 \\ \hline
NADST (N=1)                   & \checkmark      &  -      & 78.0                                                                           & 63.8                                                                       & 44.9                                                                         & 11.6                                                                                & 4.6                                                                                 & 3.2                                                                                 \\
NADST (N=1)                   & \checkmark      & \checkmark     & 78.4                                                                           & 64.0                                                                       & 47.7                                                                         & 12.7                                                                                & 6.1                                                                                 & 5.5                                                                                 \\
NADST (N=3)                   & \checkmark      & -      & 80.6                                                                           & 67.3                                                                       & 50.2                                                                         & 15.3                                                                                & 6.3                                                                                 & 4.3                                                                                 \\
NADST (N=3)                  & \checkmark      & \checkmark     & 79.0                                                                           & 65.1                                                                       & 49.2                                                                         & 13.3                                                                                & 6.3                                                                                 & 5.5                                                                                 \\ \hline
VDTN                   & \checkmark      & \checkmark     & \textbf{84.5}                                                                  & \textbf{72.8}                                                              & \textbf{60.4}                                                                & \textbf{28.0}                                                                       & \textbf{15.3}                                                                       & \textbf{13.1}                                                \\
VDTN+GPT2            & \checkmark      & \checkmark     & \textbf{85.2}                                                                  & \textbf{76.4}                                                              & \textbf{63.7}                                                                & \textbf{30.4}                                                                       & \textbf{16.8}                                                                       & \textbf{14.3}                                                \\
\hline
\end{tabular}
}
\caption{Performance (in \%) of VDTN vs. baselines on the test split of DVD-DST. \checkmark on ``Dial'' or ``Vid'' column indicates whether we use dialogue context or video context respectively.}
\label{tab:baseline_results}
\end{table}

\subsection{Results} 

\paragraph{Overall results.}
From Table \ref{tab:baseline_results}, we have the following observations:

\begin{itemize}
    \item we noted that simply using naive retrieval models such as \emph{Q-retrieval} achieved zero joint state accuracy only. 
\emph{State prior} achieved only about $15$\% and $8$\% F1 on object identities and object slots, showing that a model cannot simply rely on distribution bias of dialogue states. 
    \item The results of \emph{Object (random/all)} show that in DVD-DST, dialogues often focus on a subset of visual objects and an object perception model alone cannot perform well. 
    \item The performance gains of \emph{RNN} models show the benefits of neural network models compared to retrieval models. 
    The higher results of \emph{RNN(D)} against \emph{RNN(V)} showed the dialogue context is essential and reinforced the above observation.
    \item Comparing TRADE and UniConv, we noted that TRADE performed slightly better in component predictions, but was outperformed in joint state prediction metrics. 
    This showed the benefits of UniConv which avoids the assumptions of conditionally independent slots and learns to extract the dependencies between slot values. 
    \item Results of TRADE, UniConv, and NADST all displayed minor improvement when adding video inputs to dialogue inputs, displaying their weakness when exposed to cross-modality learning. 
    \item VDTN achieves significant performance gains and achieves the SOTA results in all component or joint prediction metrics. 
\end{itemize}

We also experimented with a version of VDTN in which the transformer network (Section \ref{subsec:mm_transformer}) was initialized from a GPT2-base model \citep{radford2019language} with a pretrained checkpoint released by HuggingFace\footnote{\url{https://huggingface.co/gpt2}}. 
Aside from using BPE to encode text sequences to match GPT2 embedding indices, we keep other components of the model the same.
VDTN+GPT2 is about $36\times$ bigger than our default VDTN model. 
As shown in Table \ref{tab:baseline_results}, the performance gains of VDTN+GPT2 indicates the benefits of large-scale language models (LMs). 
Another benefit of using pretrained GPT2 is faster training time as we observed the VDTN+GPT2 converged much earlier than training it from scratch. 
From these observations, we are excited to see more future extension of SOTA unimodal DST models \citep{lin-etal-2021-knowledge, dai-etal-2021-preview} and large pretrained LMs \citep{brown2020language, JMLR:v21:20-074}, especially ones with multimodal learning such as \citep{NEURIPS2019_c74d97b0, zhou2020unified}, to MM-DST task. 

\begin{table}[t]
\centering
\small
\resizebox{1.0\columnwidth}{!} {
\begin{tabular}{llrrr}
\hline
\multicolumn{1}{c}{\begin{tabular}[c]{@{}c@{}}Video self-\\ supervision\end{tabular}} & \multicolumn{1}{c}{Loss} & \multicolumn{1}{c}{\begin{tabular}[c]{@{}c@{}}Acc\end{tabular}} & \multicolumn{1}{c}{\begin{tabular}[c]{@{}c@{}}IoU \\@0.5\end{tabular}} & \multicolumn{1}{c}{\begin{tabular}[c]{@{}c@{}}IoU \\@0.7\end{tabular}} \\
\hline
None                                                                                  & N/A                      & 24.8                                                                            & 13.8                                                                             & 11.8                                                                             \\
\hline
$\mathcal{L}_{obj}$                                                                             & L1                       & 26.0                                                                            & 14.4                                                                             & 12.4                                                                             \\
$\mathcal{L}_{obj}$                                                                             & L2                       & 24.1                                                                            & 13.3                                                                             & 11.4                                                                             \\
\hline
$\mathcal{L}_{obj}$ (tracking)                                                                           & L1                       & 27.2                                                                            & 14.7                                                                             & 12.6                                                                             \\
$\mathcal{L}_{obj}$ (tracking)                                                                           & L2                       & 22.9                                                                            & 12.7                                                                             & 10.9                                                                             \\
\hline
$\mathcal{L}_{seg}$                                                                          & L1                       & \textbf{28.0}                                                                            & \textbf{15.3}                                                                             & \textbf{13.1}                                                                             \\
$\mathcal{L}_{seg}$                                                                          & L2                       & 27.4                                                                            & 14.7                                                                             & 12.7                                                                             \\
\hline
$\mathcal{L}_{obj} + \mathcal{L}_{seg}$                                                                       & L1                       & 23.7                                                                            & 13.0                                                                             & 11.2                                                                             \\
$\mathcal{L}_{obj} + \mathcal{L}_{seg}$                                                                       & L2                       & 24.3                                                                            & 13.4                                                                             & 11.6                \\
\hline
\end{tabular}
}
\caption{Accuracy (in \%) by self-supervised objectives.
$\mathcal{L}_{obj}$ (tracking) assumes access to oracle bounding box labels and treats the self-supervised learning task as an object tracking task.}
\label{tab:self_supervised_results}
\end{table}

\paragraph{Impacts of self-supervised video representation learning.}
From Table \ref{tab:self_supervised_results}, we noted that compared to a model trained only with the DST objective $\mathcal{L}_{dst}$, models enhanced with self-supervised video understanding objectives can improve the results.
However, we observe that L1 loss works more consistently than L2 loss in most cases.
Since L2 loss minimizes the squared differences between predicted and ground-truth values, it may be susceptible to outliers (of segment features or bounding boxes) in the dataset. 
Since we could not control these outliers, an L1 loss is more suitable. 

We also tested with $\mathcal{L}_{obj}$ (tracking), in which we used oracle bounding box labels during training, and simply passed the features of all objects to VDTN.
This modification treats the self-supervised learning task as an object tracking task in which all output representations are used to predict the ground-truth bounding box coordinates of all objects. 
Interestingly, we found $\mathcal{L}_{obj}$ (tracking) only improves the results insignificantly, as compared to the self-supervised learning objective $\mathcal{L}_{obj}$.
This indicates that our self-supervised learning tasks do not strongly depend on the availability of object boundary labels.

Finally, we found combining both segment-level and object-level self-supervision is not useful.
This is possibly due to our current masking strategy that masks object and segment features independently.
Therefore, the resulting context features might not be sufficient for recovering masked representations. 
Future work can be extended by studying a codependent masking technique to combine segment-based and object-based representation learning. 

\paragraph{Impacts of video features and time-based slots.}
Table \ref{tab:ablation_results_main} shows the results of different variants of VDTN models.
We observed that:

\begin{itemize}
    \item Segment-based learning is marginally more powerful than object-based learning. 
    \item By considering the temporal placement of objects and defining time-based slots, we noted the performance gains by ``Joint Obj State Acc'' ($\mathcal{B}$ vs. $\mathcal{B} \backslash time$). 
    The performance gains show the interesting relationships between temporal slots and discrete-only slots and the benefits of modelling both in dialogue states. 
    \item Finally, even with only object-level features $X_{bb}$, we still observed performance gains from using self-supervised loss $\mathcal{L}_{obj}$, confirming the benefits of better visual representation learning. 

\end{itemize}

\begin{table}[t]
\centering
\resizebox{1.0\columnwidth}{!} {
\begin{tabular}{lllrrr}
\hline
\multicolumn{1}{c}{\begin{tabular}[c]{@{}c@{}}Video \\ Features\end{tabular}} & \multicolumn{1}{c}{\begin{tabular}[c]{@{}c@{}}Dialogue \\ State\end{tabular}} & \multicolumn{1}{c}{\begin{tabular}[c]{@{}c@{}}Video \\loss\end{tabular}} &  
\multicolumn{1}{c}{\begin{tabular}[c]{@{}c@{}}Acc\end{tabular}} & \multicolumn{1}{c}{\begin{tabular}[c]{@{}c@{}}IoU \\ @0.5\end{tabular}} & \multicolumn{1}{c}{\begin{tabular}[c]{@{}c@{}}IoU \\@0.7\end{tabular}} \\
\hline
$X_{bb}$                                             & $\mathcal{B} \backslash \mathrm{time}$                                                                              & -                                                                                                 & 17.9                                                                             & N/A                                                                                & N/A                                                                                \\
$X_{bb}+X_{cnn}$                                          &  $\mathcal{B} \backslash \mathrm{time}$                                                                              & -                                                                      & 22.4                                                                             & N/A                                                                                & N/A                                                                                \\ \hline
$X_{bb}$                                             & $\mathcal{B}$                                                                          & -                                                                               & 19.3                                                                             & 11.0                                                                             & 9.5                                                                              \\
$X_{bb}+X_{cnn}$                                            & $\mathcal{B}$                                                                          & -                                                                                                                                             & 24.8                                                                             & 13.8                                                                             & 11.8                                                                             \\ \hline
$X_{bb}$                                             & $\mathcal{B}$                                                                          & $\mathcal{L}_{obj}$                                                                                                 & 24.0                                                                             & 12.9                                                                             & 11.0                                                                             \\
$X_{bb}+X_{cnn}$                                            & $\mathcal{B}$                                                                          & $\mathcal{L}_{obj}$                                                                                                                           & 26.0                                                                             & 14.4                                                                             & 12.4                                                                             \\
\hline
$X_{bb}+X_{cnn}$                                            & $\mathcal{B}$                                                                          & $\mathcal{L}_{seg}$                                                                                                      & \textbf{28.0}                                                                    & \textbf{15.3}                                                                    & \textbf{13.1}                   \\
\hline
\end{tabular}
}
\caption{Accuracy (in \%) by video features and state formulations}
\label{tab:ablation_results_main}
\end{table}

\begin{figure}[t]
	\centering
	\resizebox{1.0\columnwidth}{!} {
	\includegraphics{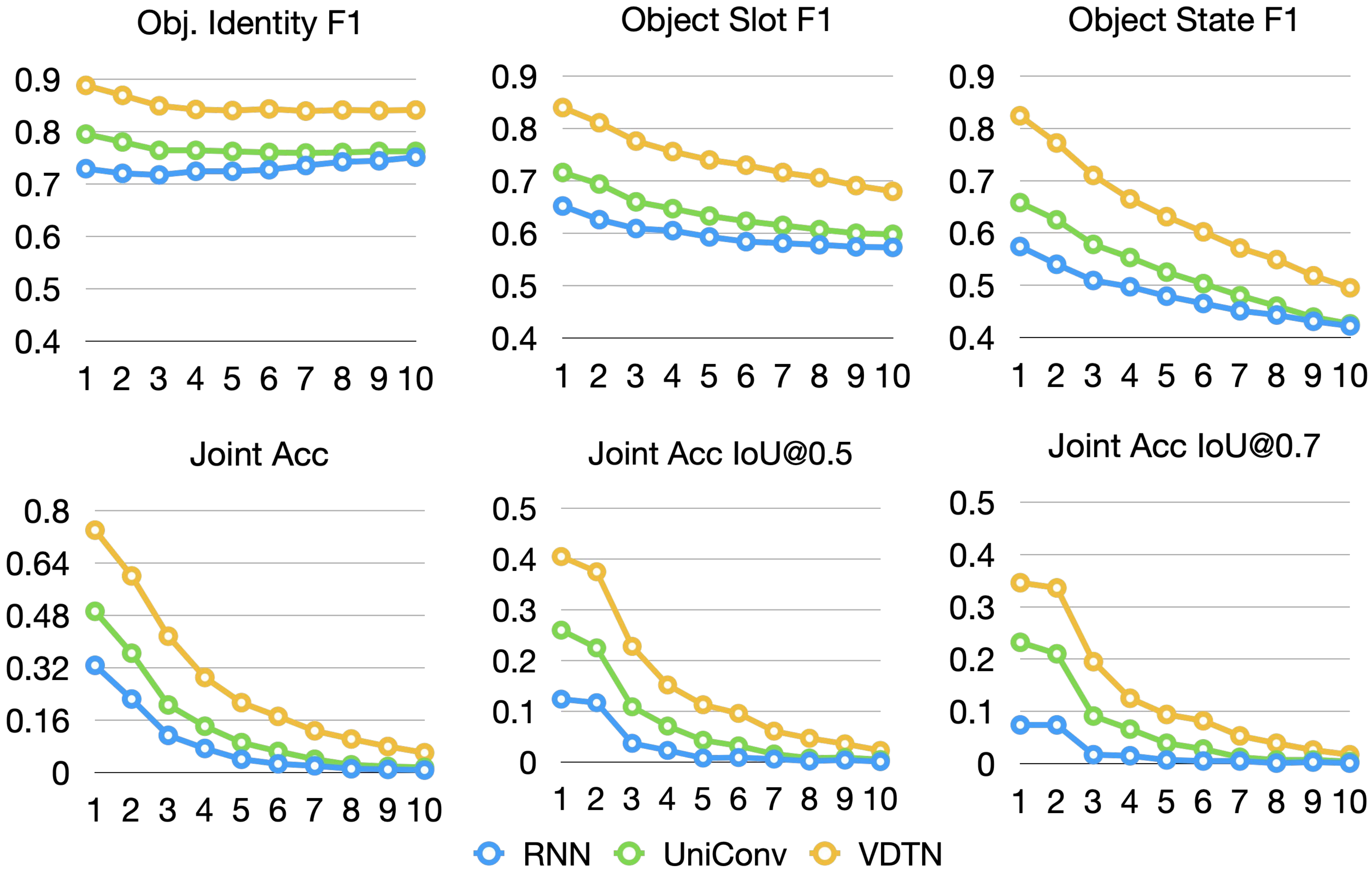}
	}
	\caption{
	Ablation results of VDTN and baselines by dialogue turn positions ($x$ axis)
	}
	\label{fig:ablation_turn_main}
\end{figure}

\paragraph{Ablation analysis by turn positions.}
Figure \ref{fig:ablation_turn_main} reported the results of VDTN predictions of states that are separated by the corresponding dialogue positions. 
The results are from the VDTN model trained with both $\mathcal{L}_{dst}$ and $\mathcal{L}_{seg}$.
As expected, we observed a downward trend of results as the turn position increases. 
We noted that state accuracy reduces more dramatically (as shown by ``Joint Acc'') than the F1 metrics of component predictions. 
For instance, ``Object Identity F1'' shows almost stable performance lines through dialogue turns. 
Interestingly, we noted that the prediction performance of dialogue states with temporal slots only deteriorates dramatically after turn $2$ onward. 
We expected that VDTN is able to learn short-term dependencies ($~1$-turn distance) between temporal slots, but failed to deal with long-term dependencies ($>1$-turn distance) between temporal slots. 
In all metrics, we observed VDTN outperforms both RNN baseline and UniConv \citep{le-etal-2020-uniconv}, across all turn positions. 
However, future work is needed to close the performance gaps between lower and higher turn positions. 

\paragraph{Impacts on downstream response prediction task.}
Finally, we tested the benefits of studying multimodal DST for a response prediction task. 
Specifically, we used the best VDTN model to predict dialogue states across all samples in DVD-DST.
We then used the predicted slots, including object identities and temporal slots, to select the video features.
The features are the visual objects and segments that are parts of the predicted dialogue states. 
We then used these selected features as input to train new Transformer decoder models which are added with an MLP as the response prediction layer.
Note that these models are trained only with a cross-entropy loss to predict answer candidates. 
From  Table \ref{tab:response_results}, we observed the benefits of filtering visual inputs by predicted states, with up to $5.9$\% accuracy score improvement \footnote{The response prediction performance is lower than the results reported by DVD \citep{le-etal-2021-dvd} as the training splits are not the same; DVD-DST has about $10$x smaller training data than \citep{le-etal-2021-dvd}.}. 
Note that there are more sophisticated approaches such as neural module networks \citep{hu2018explainable} and symbolic reasoning \citep{Chen2020Neural} to fully exploit the decoded dialogue states. 
We leave these extensions for future research. 
\begin{table}[t]
\centering
\small
\begin{tabular}{lr}
\hline
Dialogue State & \multicolumn{1}{c}{ Response Accuracy} \\ \hline
No state                   & 43.0                                \\
$\mathcal{B} \backslash \mathrm{time}$          & 46.8/47.1                         \\
$\mathcal{B}$      & 48.7/48.9                  \\ \hline      
\end{tabular}
\caption{Accuracy (in \%) of response predictions (by greedy/beam search states)}
\label{tab:response_results}
\end{table}

For more experiment results, analysis, and qualitative examples, please refer to Appendix \ref{app:results}. 
\section{Discussion and Conclusion} 
Compared to conventional DST \citep{mrksic2017neuralbelief, lei2018sequicity, gao2019dialog, Le2020Non-Autoregressive}, we show that the scope of DST can be further extended to a multimodal world.
Compared to prior work in multimodal dialogues \citep{das2017visual, hori2019avsd, thomason:corl19} which focuses more on vision-language interactions, 
our work was inspired from a dialogue-based strategy with a formulation of a dialogue state tracking task.
For more comparison to related work, please refer to Appendix \ref{app:related_work}. 

We noted the current work are limited to a synthetic benchmark with a limited video domain (3D objects).
However, we expect that MM-DST task is still applicable and can be extended to other video domains (e.g. videos of humans). 
We expect that MM-DST is useful in dialogues centered around a ``focus group'' of objects.
For further discussion of limitations, please refer to Appendix \ref{app:discuss}.

In summary, in this work, we introduced a novel MM-DST task that tracks visual objects and their attributes mentioned in dialogues.
For this task, we experimented on a synthetic benchmark with videos simulated in a 3D environment and dialogues grounded on these objects.
Finally we proposed VDTN, a Transformer-based model with self-supervised learning objectives on object and segment-level visual representations. 
\section{Broader Impacts}
During the research of this work, there is no human subject involved.
The data is used from a synthetically developed dataset, in which all videos are simulated in a 3D environment with synthetic non-human visual objects. 
We intentionally chose this dataset to minimize any distribution bias and make fair comparisons between all baseline models. 

However, we wanted to emphasize on ethical usage of any potential adaptation of our methods in real applications. Considering the development of AI in various industries, the technology introduced in this paper may be used in practical applications, such as dialogue agents with human users. 
In these cases, the adoption of the MM-DST task or VDTN should be strictly used to improve the model performance and only for legitimate and authorized purposes. 
It is crucial that any plan to apply or extend MM-DST in real systems should consider carefully all potential stakeholders as well as the risk profiles of application domains.
For instance, in case a dialogue state is extended to human subjects, any information used as slots should be clearly informed and approved by the human subjects before the slots are tracked. 

\bibliography{custom} 
\bibliographystyle{acl_natbib}

\appendix

\section{Details of Related Work}
\label{app:related_work}
Our work is related to the following domains: 

\subsection{Dialogue State Tracking}
Dialogue State Tracking (DST) research aims to develop models that can track essential information conveyed in dialogues between a dialogue agent and human (defined as hidden information state by \cite{YOUNG2010150} or belief state by \cite{mrksic2017neuralbelief}). 
DST research has evolved largely within the domain of task-oriented dialogue systems.
DST is conventionally designed in a modular dialogue system \citep{wen-etal-2017-network, budzianowski-etal-2018-multiwoz, le-etal-2020-uniconv} and preceded by a Natural Language Understanding (NLU) component. 
NLU learns to label sequences of dialogue utterances and provides a tag for each word token (often in the form of In-Out-Begin representations) \citep{kurata-etal-2016-leveraging, shi-etal-2016-recurrent, rastogi2017ScalableMD}. 
To avoid credit assignment problems and streamline the modular designs, NLU and DST have been integrated into a single module \citep{mrksic2017neuralbelief, xu2018end, zhong-etal-2018-global}. 
These DST approaches can be roughly categorized into two types: fixed-vocabulary or open-vocabulary. 
Fixed-vocabulary approaches are designed for classification tasks which assume a fixed set of \emph{(slot, value)} candidates and directly retrieve items from this set to form dialogue states during test time \citep{henderson2014RobustDS, ramadan2018large, lee2019sumbt}.
More recently, we saw more approaches toward open-vocabulary strategies which learn to generate candidates based on input dialogue context \citep{lei2018sequicity, gao2019dialog, wu-etal-2019-transferable, Le2020Non-Autoregressive}. 
Our work is more related to open-vocabulary DST, but we essentially redefined the DST task with multimodality. 
Based on our literature review, we are the first to formally extend DST and bridge the gap between traditional task-oriented dialogues and multimodal dialogues. 

\subsection{Visually-grounded Dialogues}

A novel challenge to machine intelligence,  the intersection of vision and language research has expanded considerably in the past few years. 
Earlier benchmarks test machines to perceive visual inputs,  and learn to generate captions \citep{farhadi2010every, lin2014microsoft, rohrbach2015dataset}, ground text phrases and objects \citep{kazemzadeh2014referitgame, plummer2015flickr30k}, and answer questions about the visual contents \citep{antol2015vqa, zhu2016visual7w, jang2017tgif, lei-etal-2018-tvqa}. 
As an orthogonal development from Visual Question Answering problems, we noted recent work that targets vision-language in dialogue context, in which an image or video is given and the dialogue utterances are centered around its visual contents \citep{de2017guesswhat, das2017visual, visdial_eval, hori2019avsd, thomason:corl19, le-etal-2021-dvd}. 
Recent work has addressed different challenges in visually-grounded dialogues, including multimodal integration \citep{hori2019avsd, le-etal-2019-multimodal, 9376902}, cross-turn dependencies \citep{das2017learning, schwartz2019factor, le2021learning}, visual understanding \citep{le-etal-2020-bist}, and data distribution bias \citep{qi2020two}. 
Our work is more related to the challenge of visual object reasoning \citep{seo2017visual, kottur2018visual}, but focused on a multi-turn tracking task over multiple turns of dialogue context. 
The prior approaches are not well designed to track objects and maintain a recurring memory or state of these objects from turn to turn. 
This challenge becomes more obvious when a dialogue involves multiple objects of similar characters or appearance. 
We directly tackles this challenge as we formulated a novel multimodal state tracking task and leveraged the research development from DST in task-oriented dialogue systems. 
As shown in our experiments, baseline models that use attention strategies similar to \citep{seo2017visual, kottur2018visual} did not perform well in MM-DST. 

\subsection{Multimodal DST}

\begin{table*}[htbp]
\centering
\small
\begin{tabular}{lllll}
\hline
\multicolumn{1}{c}{\textbf{Split}} & \multicolumn{1}{c}{\textbf{\# Videos}} & \multicolumn{1}{c}{\textbf{\# Dialogues}} & \multicolumn{1}{c}{\textbf{\# Turns}} & \multicolumn{1}{c}{\textbf{\# Slots}} \\
\hline
DVD-DST-Train                              & 9300                                   & 9295                                      & 92950   & 6                                    \\
DVD-DST-Val                         & 3327                                   & 3326                                      & 33260       & 6                                \\
DVD-DST-Test                               & 1371                                   & 1371                                      & 13710        & 6       \\
DVD-DST-All                               & 13998                                   & 13992                                      & 139920        & 6       \\
\hline
MultiWOZ \citep{budzianowski-etal-2018-multiwoz}                              & N/A                                   & 8438                                      & 115424   & 25                                    \\
CarAssistant \citep{eric-etal-2017-key}                              & N/A                                   &      2425                                 & 12732   &                           13          \\

WOZ2 \citep{wen-etal-2017-network}                              & N/A                                   &      600                                 & 4472   &                           4          \\
DSTC2 \citep{henderson2014second}                              & N/A                                   &      1612                                 & 23354   &                           8          \\

\hline
\end{tabular}
\caption{Dataset summary: statistics of related benchmarks are from \citep{budzianowski-etal-2018-multiwoz}}
\label{tab:data}
\end{table*}

\begin{figure*}[htbp]
	\centering
	\resizebox{0.8\textwidth}{!} {
	\includegraphics{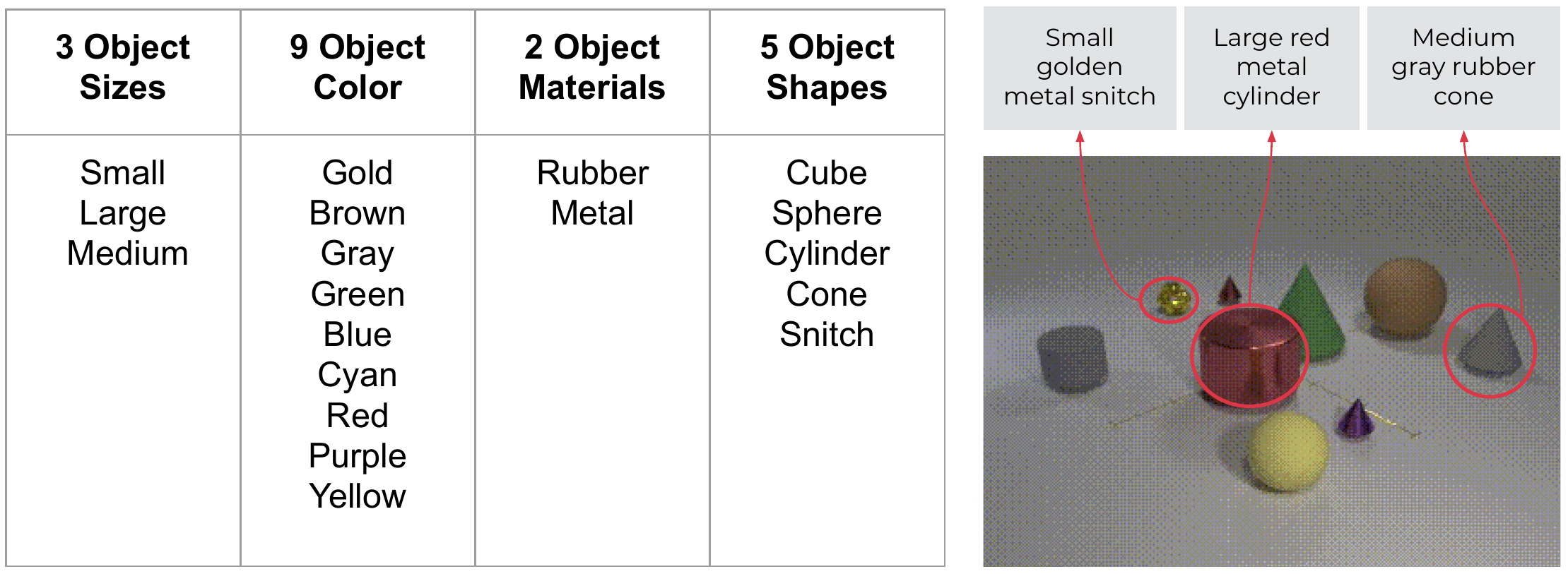}
	}
	\caption{
	Synthetic visual objects in the CATER universe 
	}
	\label{fig:cater}
	\vspace{-0.1in}
\end{figure*}

We noted a few studies have attempted to integrate some forms of state tracking in multimodal dialogues. 
In \citep{mou2020multimodal}, however, we are not convinced that a dialogue state tracking task is a major focus, or correctly defined. 
In \citep{pang2020visual}, we noted that some form of object tracking is introduced throughout dialogue turns.
The tracking module is used to decide which object the dialogue centers around.
This method extends to multi-object tracking but the objects are only limited within static images, and there is no recurring information state (object attributes) maintained at each turn. 
Compared to our work, their tracking module only requires object identity as a single-slot state from turn to turn.
Almost concurrent to our work, we noted \citep{kottur-etal-2021-simmc} which formally, though very briefly, focuses on multimodal DST. However, the work is limited to the task-oriented domain, and each dialogue is only limited to a single goal-driven object in a synthetic image. While this definition is useful in the task-oriented dialogue domain, it does not account for the DST of multiple visual objects as defined in our work. 

\section{DVD-DST Dataset Details}
\label{app:data}

For each of CATER videos from the extended split \citep{shamsian2020learning}, we generated up to 10 turns for each CATER video.
In total, DVD-DST contains more than $13k$ dialogues, resulting in more $130k$ (human, system) utterance pairs and corresponding dialogue states.
A comparison of statistics of DVD-DST and prior DST benchmarks can be seen in Table \ref{tab:data}. 
We observed that DVD-DST contains a larger scale data than the related DST benchmark.
Even though the number of slots in DVD-DST is only $6$, lower than prior state tracking datasets, our experiments indicate that most current conventional DST models perform poorly on DVD-DST.

\textbf{CATER universe.}
Figure \ref{fig:cater} displays the configuration of visual objects in the CATER universe. 
In total, there are 3 object sizes, 9 colors, 2 materials, and 5 shapes. 
These attributes are combined randomly to synthesize objects in each CATER video. 
We directly adopted these attributes as slots in dialogue states, and each dialogue utterance frequently refers to these objects by one or more attributes. 
In total, there are 193 (size, color, material, shape) valid combinations, each of which corresponds to an object class in our models. 

\textbf{Sample dialogues.} Please refer to Figure \ref{fig:002660}, Table \ref{tab:example_002660} and Table \ref{tab:example_001441}.

\textbf{Usage.}
We want to highlight that the DVD-DST dataset should only be used for its intended purpose, i.e. to diagnose  dialogue systems on their tracking abilities. 
Any derivatives of the data should be limited within the research contexts of MM-DST. 

\section{Additional Training Details}
\label{app:training}
In practice, we applied label smoothing \citep{szegedy2016rethinking} on state sequence labels to regularize the training. 
As the segment-level representations are stacked by the number of objects, we randomly selected only one vector per masked segment to apply $\mathcal{L}_{seg}$.
We tested both L1 and L2 losses on $\mathcal{L}_{bb/seg}$.
All model parameters, except pretrained visual perception models, are initialized by a uniform distribution \citep{glorot2010understanding}. 

\begin{table*}[h]
\centering
\resizebox{1.0\textwidth}{!} {
\begin{tabular}{lllrrrrrr}
\hline
\multicolumn{1}{c}{\multirow{2}{*}{\begin{tabular}[c]{@{}c@{}}Video \\ Features\end{tabular}}} & \multicolumn{1}{c}{\multirow{2}{*}{\begin{tabular}[c]{@{}c@{}}Dialogue \\ State\end{tabular}}} & \multicolumn{1}{c}{\multirow{2}{*}{\begin{tabular}[c]{@{}c@{}}Video\\loss\end{tabular}}} & \multicolumn{3}{c}{Greedy}                                                                                                                                                                                                                          & \multicolumn{3}{c}{Beam Search}                                                                                                                                                                                                                              \\
\cline{4-6}
\cline{7-9}
\multicolumn{1}{c}{}                                & \multicolumn{1}{c}{}                                                                           & \multicolumn{1}{c}{}                                                                                      & \multicolumn{1}{c}{\begin{tabular}[c]{@{}c@{}}Joint Obj \\ State Acc\end{tabular}} & \multicolumn{1}{c}{\begin{tabular}[c]{@{}c@{}}Joint State \\ IoU@0.5\end{tabular}} & \multicolumn{1}{c}{\begin{tabular}[c]{@{}c@{}}Joint State \\ IoU@0.7\end{tabular}} & \multicolumn{1}{c}{\begin{tabular}[c]{@{}c@{}}Joint Obj \\ State Acc\end{tabular}} & \multicolumn{1}{c}{\begin{tabular}[c]{@{}c@{}}Joint State \\ IoU@0.5\end{tabular}} & \multicolumn{1}{c}{\begin{tabular}[c]{@{}c@{}}Joint State \\ IoU@0.7\end{tabular}} \\
\hline
$X_{bb}$                                             & $\mathcal{B} \backslash \mathrm{time}$                                                                              & -                                                                                                      & 17.3\%                                                                             & N/A                                                                                & N/A                                                                                & 17.9\%                                                                             & N/A                                                                                & N/A                                                                                \\
$X_{bb}+X_{cnn}$                                          &  $\mathcal{B} \backslash \mathrm{time}$                                                                              & -                                                                                                      & 20.0\%                                                                             & N/A                                                                                & N/A                                                                                & 22.4\%                                                                             & N/A                                                                                & N/A                                                                                \\ \hline
$X_{bb}$                                             & $\mathcal{B}$                                                                          & -                                                                                                      & 16.6\%                                                                             & 9.6\%                                                                              & 8.3\%                                                                              & 19.3\%                                                                             & 11.0\%                                                                             & 9.5\%                                                                              \\
$X_{bb}+X_{cnn}$                                            & $\mathcal{B}$                                                                          & -                                                                                                      & 22.4\%                                                                             & 12.7\%                                                                             & 10.8\%                                                                             & 24.8\%                                                                             & 13.8\%                                                                             & 11.8\%                                                                             \\ \hline
$X_{bb}$                                             & $\mathcal{B}$                                                                          & $\mathcal{L}_{obj}$                                                                                                        & 21.7\%                                                                             & 11.7\%                                                                             & 10.0\%                                                                             & 24.0\%                                                                             & 12.9\%                                                                             & 11.0\%                                                                             \\
$X_{bb}+X_{cnn}$                                            & $\mathcal{B}$                                                                          & $\mathcal{L}_{obj}$                                                                                                        & 23.1\%                                                                             & 13.2\%                                                                             & 11.3\%                                                                             & 26.0\%                                                                             & 14.4\%                                                                             & 12.4\%                                                                             \\
\hline
$X_{bb}+X_{cnn}$                                            & $\mathcal{B}$                                                                          & $\mathcal{L}_{seg}$                                                                                                     & \textbf{24.3\%}                                                                    & \textbf{13.4\%}                                                                    & \textbf{11.4\%}                                                                    & \textbf{28.0\%}                                                                    & \textbf{15.3\%}                                                                    & \textbf{13.1\%}                   \\
\hline
\end{tabular}
}
\caption{Ablation results by joint state predictions, using greedy or beam search decoding styles}
\label{tab:ablation_results}
\end{table*}
\begin{table*}[h]
\centering
\resizebox{1.0\textwidth}{!} {
\begin{tabular}{lllrrrrrrr}
\hline
\multicolumn{1}{c}{\begin{tabular}[c]{@{}c@{}}Video \\ Features\end{tabular}} & \multicolumn{1}{c}{\begin{tabular}[c]{@{}c@{}}Dialogue\\ State\end{tabular}} & \multicolumn{1}{c}{\begin{tabular}[c]{@{}c@{}}Video \\ self-\\ supervision\end{tabular}} & \multicolumn{1}{c}{\begin{tabular}[c]{@{}c@{}}Obj \\ Identity \\ F1\end{tabular}} & \multicolumn{1}{c}{\begin{tabular}[c]{@{}c@{}}Obj \\ Slot \\ F1\end{tabular}} & \multicolumn{1}{c}{\begin{tabular}[c]{@{}c@{}}Obj \\ State \\ F1\end{tabular}} & \multicolumn{1}{c}{\begin{tabular}[c]{@{}c@{}}Size \\ F1\end{tabular}} & \multicolumn{1}{c}{\begin{tabular}[c]{@{}c@{}}Color\\  F1\end{tabular}} & \multicolumn{1}{c}{\begin{tabular}[c]{@{}c@{}}Material \\ F1\end{tabular}} & \multicolumn{1}{c}{\begin{tabular}[c]{@{}c@{}}Shape\\  F1\end{tabular}} \\
\hline
$X_{bb}$                                                                       & $\mathcal{B} \backslash \mathrm{time}$                                                            & -                                                                                     & 79.4\%                                                                            & 64.2\%                                                                        & 48.5\%                                                                         & 55.9\%                                                                 & 76.6\%                                                                  & 41.4\%                                                                     & 63.5\%                                                                  \\
$X_{bb}+X_{cnn}$                                                                      & $\mathcal{B} \backslash \mathrm{time}$                                                            & -                                                                                     & 81.4\%                                                                            & 66.9\%                                                                        & 52.5\%                                                                         & 58.0\%                                                                 & 79.4\%                                                                  & 39.5\%                                                                     & 66.6\%                                                                  \\
\hline
$X_{bb}$                                                                       & $\mathcal{B}$                                                         & -                                                                                     & 78.5\%                                                                            & 63.6\%                                                                        & 49.8\%                                                                         & 56.5\%                                                                 & 76.4\%                                                                  & 38.8\%                                                                     & 63.1\%                                                                  \\
$X_{bb}+X_{cnn}$                                                                      & $\mathcal{B}$                                                         & -                                                                                     & 83.3\%                                                                            & 69.4\%                                                                        & 55.1\%                                                                         & 56.7\%                                                                 & 81.8\%                                                                  & 47.0\%                                                                     & 69.8\%                                                                  \\
\hline
$X_{bb}$                                                                       & $\mathcal{B}$                                                         & $\mathcal{L}_{obj}$                                                                                & 82.2\%                                                                            & 69.5\%                                                                        & 56.2\%                                                                         & 61.4\%                                                                 & 81.0\%                                                                  & 44.9\%                                                                     & 69.9\%                                                                  \\
$X_{bb}+X_{cnn}$                                                                      & $\mathcal{B}$                                                         & $\mathcal{L}_{obj}$                                                                                & \textbf{84.7\%}                                                                   & 72.0\%                                                                        & 58.6\%                                                                         & 59.7\%                                                                 & 83.5\%                                                                  & \textbf{52.3\%}                                                            & 71.7\%                                                                  \\
\hline
$X_{bb}+X_{cnn}$                                                                      & $\mathcal{B}$                                                         & $\mathcal{L}_{seg}$                                                                             & 84.5\%                                                                            & \textbf{72.8\%}                                                               & \textbf{60.4\%}                                                                & \textbf{64.1\%}                                                        & \textbf{84.2\%}                                                         & 50.9\%                                                                     & \textbf{71.9\%}                                 \\
\hline
\end{tabular}
}
\caption{Ablation results by component predictions of object identities, slots, and object states}
\label{tab:ablation_results_f1}
\end{table*}

For fair comparison among baselines, all models use both object-level and segment-level feature representations, encoded by the same method as Describe in Section \ref{subsec:video_encode}.
In \emph{TRADE}, the video representations are passed to an RNN encoder, and the output hidden states are concatenated to the dialogue hidden states. 
Both are passed to the original pointer-based decoder. 
In \emph{UniConv} and \emph{NADAST}, we stacked another Transformer attention layer to attend on video representations before the original state-to-dialogue attention layer. 
We all baseline models, we replaced the original (domain, slot) embeddings as (object class, slot) embeddings and kept the original model designs.

Note that in our visual perception model, we adopted the finetuned Faster R-CNN model used by \cite{shamsian2020learning}.
The model was finetuned to predict object bounding boxes and object classes. 
The object classes are derived based on object appearance, based on the four attributes of size, color, material, and shape. 
In total, there are 193 object classes. 
For segment embeddings, we adopted the ResNeXt-101 model \citep{xie2017aggregated} finetuned on Kinetics dataset \citep{kay2017kinetics}.  
For all models (except for VDTN ablation analysis), we standardized $N_{obj}=10$ and $N_{stride}=12$ to sub-sample object and segment-level embeddings. 

\textbf{Resources.}
Note that all experiments did not require particularly large computing resources as we limited all model training to a single GPU, specifically on a Tesla V100 GPU of 16G configuration. 

\section{Additional Results}
\label{app:results}

\textbf{Greedy vs. Beam Search Decoding.}
Table \ref{tab:ablation_results} shows the results of different variants of VDTN models.
We observed that compared to greedy decoding, beam search decoding improves the performance in all models.
As beam search decoding selects the best decoded state by the joint probabilities of tokens, this observation indicates the benefits of considering slot values to be co-dependent and their relationships should be modelled.
This is consistent with similar observations in later work of unimodal DST \citep{lei2018sequicity, Le2020Non-Autoregressive}. 

\begin{table*}[htbp]
\centering
\resizebox{1.0\textwidth}{!} {
\begin{tabular}{lllrrrrrr}
\hline
\multicolumn{1}{c}{Video Features} & \multicolumn{1}{c}{\begin{tabular}[c]{@{}c@{}}Dialogue \\ State\end{tabular}} & \multicolumn{1}{c}{\begin{tabular}[c]{@{}c@{}}Video self-\\ supervision\end{tabular}} & \multicolumn{1}{c}{\begin{tabular}[c]{@{}c@{}}Obj Identity \\ Recall\end{tabular}} & \multicolumn{1}{c}{\begin{tabular}[c]{@{}c@{}}Obj Identity \\ Precision\end{tabular}} & \multicolumn{1}{c}{\begin{tabular}[c]{@{}c@{}}Obj Slot \\ Recall\end{tabular}} & \multicolumn{1}{c}{\begin{tabular}[c]{@{}c@{}}Obj Slot \\ Precision\end{tabular}} & \multicolumn{1}{c}{\begin{tabular}[c]{@{}c@{}}Obj State \\ Recall\end{tabular}} & \multicolumn{1}{c}{\begin{tabular}[c]{@{}c@{}}Obj State \\ Precision\end{tabular}} \\
\hline
$X_{bb}$                            & $\mathcal{B} \backslash \mathrm{time}$                                                             & -                                                                                  & 77.2\%                                                                             & 81.8\%                                                                                & 65.0\%                                                                         & 63.4\%                                                                            & 47.1\%                                                                          & 50.0\%                                                                             \\
$X_{bb}+X_{cnn}$                           & $\mathcal{B} \backslash \mathrm{time}$                                                             & -                                                                                  & 75.1\%                                                                             & 88.8\%                                                                                & 63.1\%                                                                         & 71.3\%                                                                            & 48.5\%                                                                          & 57.3\%                                                                             \\
\hline
$X_{bb}$                            & $\mathcal{B}$                                                         & -                                                                                  & 73.6\%                                                                             & 84.1\%                                                                                & 61.7\%                                                                         & 65.7\%                                                                            & 46.7\%                                                                          & 53.4\%                                                                             \\
$X_{bb}+X_{cnn}$                           & $\mathcal{B}$                                                         & -                                                                                  & 78.2\%                                                                             & 89.1\%                                                                                & 66.2\%                                                                         & 73.0\%                                                                            & 51.7\%                                                                          & 58.9\%                                                                             \\
\hline
$X_{bb}$                            & $\mathcal{B}$                                                         & $\mathcal{L}_{obj}$                                                                                    & 76.4\%                                                                             & 88.9\%                                                                                & 67.4\%                                                                         & 71.7\%                                                                            & 52.2\%                                                                          & 60.8\%                                                                             \\
$X_{bb}+X_{cnn}$                           & $\mathcal{B}$                                                         & $\mathcal{L}_{obj}$                                                                                    & 80.1\%                                                                             & \textbf{90.0\%}                                                                       & 69.1\%                                                                         & 75.2\%                                                                            & 55.4\%                                                                          & 62.2\%                                                                             \\
\hline
$X_{bb}+X_{cnn}$                           & $\mathcal{B}$                                                         & $\mathcal{L}_{seg}$                                                                                 & \textbf{80.5\%}                                                                    & 89.0\%                                                                                & \textbf{70.2\%}                                                                & \textbf{75.6\%}                                                                   & \textbf{57.6\%}                                                                 & \textbf{63.6\%}                                                                 \\
\hline
\end{tabular}
}
\caption{Ablation results by individual object identity/slot/state}
\label{tab:ablation_results_obj}
\end{table*}

\begin{table*}[h]
\centering
\resizebox{1.0\textwidth}{!} {
\begin{tabular}{lllrrrrrrrr}
\hline
\multicolumn{1}{c}{Video Features} & \multicolumn{1}{c}{\begin{tabular}[c]{@{}c@{}}Dialogue\\  State\end{tabular}} & \multicolumn{1}{c}{\begin{tabular}[c]{@{}c@{}}Video self-\\ supervision\end{tabular}} & \multicolumn{1}{c}{\begin{tabular}[c]{@{}c@{}}Size \\ Recall\end{tabular}} & \multicolumn{1}{c}{\begin{tabular}[c]{@{}c@{}}Size \\ Precision\end{tabular}} & \multicolumn{1}{c}{\begin{tabular}[c]{@{}c@{}}Color \\ Recall\end{tabular}} & \multicolumn{1}{c}{\begin{tabular}[c]{@{}c@{}}Color \\ Precision\end{tabular}} & \multicolumn{1}{c}{\begin{tabular}[c]{@{}c@{}}Material \\ Recall\end{tabular}} & \multicolumn{1}{c}{\begin{tabular}[c]{@{}c@{}}Material\\ Precision\end{tabular}} & \multicolumn{1}{c}{\begin{tabular}[c]{@{}c@{}}Shape \\ Recall\end{tabular}} & \multicolumn{1}{c}{\begin{tabular}[c]{@{}c@{}}Shape \\ Precision\end{tabular}} \\
\hline
$X_{bb}$                            & $\mathcal{B} \backslash \mathrm{time}$                                                             & -                                                                                  & 60.1\%                                                                     & 52.2\%                                                                        & 76.8\%                                                                      & 76.4\%                                                                         & 43.2\%                                                                         & 39.7\%                                                                           & 61.4\%                                                                      & 65.6\%                                                                         \\
$X_{bb}+X_{cnn}$                           & $\mathcal{B} \backslash \mathrm{time}$                                                             & -                                                                                  & 52.0\%                                                                     & 65.6\%                                                                        & 76.2\%                                                                      & 82.9\%                                                                         & 34.8\%                                                                         & 45.8\%                                                                           & 65.5\%                                                                      & 67.8\%                                                                         \\
\hline
$X_{bb}$                            & $\mathcal{B}$                                                         & -                                                                                  & 52.0\%                                                                     & 61.9\%                                                                        & 72.0\%                                                                      & 81.2\%                                                                         & 40.8\%                                                                         & 37.1\%                                                                           & 63.3\%                                                                      & 63.0\%                                                                         \\
$X_{bb}+X_{cnn}$                           & $\mathcal{B}$                                                         & -                                                                                  & 49.4\%                                                                     & 66.5\%                                                                        & 79.2\%                                                                      & 84.6\%                                                                         & 45.0\%                                                                         & 49.2\%                                                                           & 68.9\%                                                                      & 70.6\%                                                                         \\
\hline
$X_{bb}$                            & $\mathcal{B}$                                                         & $\mathcal{L}_{obj}$                                                                                    & 59.6\%                                                                     & 63.4\%                                                                        & 79.3\%                                                                      & 82.9\%                                                                         & 43.8\%                                                                         & 46.0\%                                                                           & 66.6\%                                                                      & 73.5\%                                                                         \\
$X_{bb}+X_{cnn}$                           & $\mathcal{B}$                                                         & $\mathcal{L}_{obj}$                                                                                    & 54.1\%                                                                     & 66.6\%                                                                        & 82.4\%                                                                      & 84.7\%                                                                         & \textbf{48.8\%}                                                                & \textbf{56.3\%}                                                                  & \textbf{69.3\%}                                                             & 74.3\%                                                                         \\ \hline
$X_{bb}+X_{cnn}$                           & $\mathcal{B}$                                                         & $\mathcal{L}_{seg}$                                                                                 & \textbf{60.9\%}                                                            & \textbf{67.7\%}                                                               & \textbf{83.2\%}                                                             & \textbf{85.4\%}                                                                & 48.6\%                                                                         & 53.4\%                                                                           & 67.9\%                                                                      & \textbf{76.5\%}                                                                                                                                     \\ \hline
\end{tabular}
}
\caption{Ablation results by individual slot type}
\label{tab:ablation_results_slot}
\end{table*}

\paragraph{Ablation analysis by component predictions.}
From Table \ref{tab:ablation_results_f1}, we have the following observations: 
(1) In ablation results by component predictions, we noted that models can generally detect object identities well with F1 about $80$\%. 
However, when considering object and slot tuples, F1 reduces to $48-60$\%, indicating the gaps are caused by slot value predictions. 
(2) By individual slots, we noted ``color'' and ``shape'' slots are easier to track than ``size'' and ``material'' slots. 
We noted that in the CATER universe, the latter two slots have lower visual variances (less possible values) than the others.
As a result, objects are more likely to share the same size or material and hence, discerning objects by those slots and tracking them in dialogues become more challenging. 

Table \ref{tab:ablation_results_obj} and \ref{tab:ablation_results_slot} display the ablation results by component predictions, using precision and recall metrics.
We still noted consistent observations as described in Section \ref{sec:exps}.
Notably, we found that current VDTN models are better in tuning the correct predictions (as shown by high precision metrics) but still fail to select all components as a set (as shown by low recall metrics). 
This might be caused by the upstream errors coming from the visual perception models, which may fail to visually perceive all objects and their attributes. 

\paragraph{Results by turn positions.}
Table \ref{tab:ablation_turn} reported the results of VDTN predictions of states that are separated by the corresponding dialogue positions. 
The results are from the VDTN model trained with both $\mathcal{L}_{dst}$ and $\mathcal{L}_{seg}$.
As expected, we observed a downward trend of results as the turn position increases. 

\begin{table*}[h]
\centering
\small
\begin{tabular}{crrrrrr}
\hline
\begin{tabular}[c]{@{}c@{}}Turn \\ Position\end{tabular} & \multicolumn{1}{c}{\begin{tabular}[c]{@{}c@{}}Obj Identity \\ F1\end{tabular}} & \multicolumn{1}{c}{\begin{tabular}[c]{@{}c@{}}Obj Slot \\ F1\end{tabular}} & \multicolumn{1}{c}{\begin{tabular}[c]{@{}c@{}}Obj State\\  F1\end{tabular}} & \multicolumn{1}{c}{\begin{tabular}[c]{@{}c@{}}Joint Obj \\ State Acc\end{tabular}} & \multicolumn{1}{c}{\begin{tabular}[c]{@{}c@{}}Joint State \\ IoU@0.5\end{tabular}} & \multicolumn{1}{c}{\begin{tabular}[c]{@{}c@{}}Joint State \\ IoU@0.7\end{tabular}} \\\hline
1                                                        & \textbf{88.8\%}                                                                & \textbf{84.0\%}                                                            & \textbf{82.4\%}                                                             & \textbf{74.0\%}                                                                    & \textbf{40.5\%}                                                                    & \textbf{34.6\%}                                                                    \\
2                                                        & 86.9\%                                                                         & 81.1\%                                                                     & 77.2\%                                                                      & 60.0\%                                                                             & 37.5\%                                                                             & 33.6\%                                                                             \\
3                                                        & 84.9\%                                                                         & 77.6\%                                                                     & 71.0\%                                                                      & 41.6\%                                                                             & 22.8\%                                                                             & 19.5\%                                                                             \\
4                                                        & 84.2\%                                                                         & 75.6\%                                                                     & 66.5\%                                                                      & 29.0\%                                                                             & 15.2\%                                                                             & 12.5\%                                                                             \\
5                                                        & 84.0\%                                                                         & 74.0\%                                                                     & 63.1\%                                                                      & 21.3\%                                                                             & 11.3\%                                                                             & 9.4\%                                                                              \\
6                                                        & 84.3\%                                                                         & 73.0\%                                                                     & 60.2\%                                                                      & 17.1\%                                                                             & 9.6\%                                                                              & 8.2\%                                                                              \\
7                                                        & 83.9\%                                                                         & 71.6\%                                                                     & 57.1\%                                                                      & 12.7\%                                                                             & 6.1\%                                                                              & 5.3\%                                                                              \\
8                                                        & 84.1\%                                                                         & 70.6\%                                                                     & 54.9\%                                                                      & 10.2\%                                                                             & 4.7\%                                                                              & 3.9\%                                                                              \\
9                                                        & 84.0\%                                                                         & 69.1\%                                                                     & 51.8\%                                                                      & 7.9\%                                                                              & 3.6\%                                                                              & 2.6\%                                                                              \\
10                                                       & 84.1\%                                                                         & 68.0\%                                                                     & 49.5\%                                                                      & 6.0\%                                                                              & 2.3\%                                                                              & 1.7\%                                                                              \\\hline
Average                                                  & 84.9\%                                                                         & 74.5\%                                                                     & 63.4\%                                                                      & 28.0\%                                                                             & 15.3\%                                                                             & 13.1\%               \\ \hline                                                             
\end{tabular}
\caption{Ablation results by dialogue turn positions}
\label{tab:ablation_turn}
\end{table*}

\begin{table*}[h]
\begin{subtable}[c]{0.49\textwidth}
\centering
\small
\resizebox{0.9\textwidth}{!} {
\begin{tabular}{ccrrr}
\hline
\begin{tabular}[c]{@{}c@{}}$\mathcal{B}_{t-1}$\end{tabular} & \begin{tabular}[c]{@{}c@{}}Max\\ turns\end{tabular} & \multicolumn{1}{c}{\begin{tabular}[c]{@{}c@{}}Joint Obj\\ State Acc\end{tabular}} & \multicolumn{1}{c}{\begin{tabular}[c]{@{}c@{}}Joint State \\ IoU@0.5\end{tabular}} & \multicolumn{1}{c}{\begin{tabular}[c]{@{}c@{}}Joint State \\ IoU@0.7\end{tabular}} \\
\hline
\checkmark
                                                    & 10                                                  & 22.5\%                                                                            & 11.5\%                                                                             & 10.1\%                                                                             \\
\checkmark
                                                    & 7                                                   & 22.0\%                                                                            & 11.8\%                                                                             & 10.4\%                                                                             \\
\checkmark
                                                    & 1                                                   & {24.8\%}                                                                   & {13.8\%}                                                                    & {11.8\%}                                                                    \\
\checkmark
                                                    & 0                                                   & 22.3\%                                                                            & 12.3\%                                                                             & 10.5\%                                                                             \\
                                \hline
-                                                     & 10                                                  & 18.5\%                                                                            & 9.4\%                                                                              & 8.6\%                                                                              \\
     -                                                & 7                                                   & 19.0\%                                                                            & 9.5\%                                                                              & 8.7\%                                                                              \\
  -                                                   & 1                                                   & 7.8\%                                                                             & 4.5\%                                                                              & 4.1\%                                                                              \\
 -                                                    & 0                                                   & 1.3\%                                                                             & 0.7\%                                                                              & 0.7\%                                          \\
                                                     \hline
          \checkmark *                                                                                                & 1                                                   &  29.3\%                                                                          & 18.6\%                                                                              &    16.4\%                                      \\
\hline
\end{tabular}
}
\subcaption{dialogue encoding by prior states and dialogue sizes: \\$*$ denotes using oracle values.}
\label{tab:dialogue_results}
\end{subtable}
\begin{subtable}[c]{0.49\textwidth}
\centering
\small
\resizebox{0.96\textwidth}{!} {
\begin{tabular}{ccrrr}
\hline
\begin{tabular}[c]{@{}c@{}}$N_{obj}$\end{tabular} & \begin{tabular}[c]{@{}c@{}}$N_{stride}$\end{tabular} & \multicolumn{1}{c}{\begin{tabular}[c]{@{}c@{}}Joint Object \\ State Acc\end{tabular}} & \multicolumn{1}{c}{\begin{tabular}[c]{@{}c@{}}Joint State \\ IoU@0.5\end{tabular}} & \multicolumn{1}{c}{\begin{tabular}[c]{@{}c@{}}Joint State \\ IoU@0.7\end{tabular}} \\
\hline
10                                                 & 12                                                   & 24.8\%                                                                                & 13.8\%                                                                             & 11.8\%                                                                             \\
7                                                  & 12                                                   & 18.0\%                                                                                & 10.1\%                                                                             & 9.0\%                                                                              \\
3                                                  & 12                                                   & 4.9\%                                                                                 & 2.9\%                                                                              & 2.6\%                                                                              \\
0                                                  & 12                                                   & 1.5\%                                                                                 & 0.7\%                                                                              & 0.7\%                                                                              \\
\hline
10                                                 & 300                                                   & 28.2\%                                                                                & 6.0\%                                                                              & 3.7\%                                                                              \\

10                                                 & 24                                                   & 27.8\%                                                                                & 14.8\%                                                                              & 12.6\%                                                                              \\
10                                                 & 15                                                   & 26.3\%                                                                                & 14.4\%                                                                             & 12.4\%                                                                             \\
10                                                 & 12                                                   & 24.8\%                                                                                & 13.8\%                                                                             & 11.8\%                                                                             \\
\hline
10*                                                & 12                                                   & {29.2\%}                                                                                & {15.6\%}                                                                             & {13.4\%}                                                              \\
\hline
\end{tabular}
}
\subcaption{video encoding by number of objects and sampling strides: $*$ denotes perfect object perception.}
\label{tab:video_results}
\end{subtable}
\caption{Ablation results by encoding strategies: All models are trained only with $\mathcal{L}_{dst}$.}
\end{table*}

\begin{table*}[htbp]
\centering
\small
\begin{tabular}{lrrrrrr}
\hline
                     & \multicolumn{3}{c}{$X_{bb}+X_{cnn}$}                                                                                                                                                                                                                                & \multicolumn{3}{c}{$X_{cnn}$ only}                                                                                                                                                                                                                              \\
                     \cline{2-4} \cline{5-7}
Model                & \multicolumn{1}{c}{\begin{tabular}[c]{@{}c@{}}Joint Obj\\ State Acc\end{tabular}} & \multicolumn{1}{c}{\begin{tabular}[c]{@{}c@{}}Joint State \\ IoU@0.5\end{tabular}} & \multicolumn{1}{c}{\begin{tabular}[c]{@{}c@{}}Joint State \\ IoU@0.7\end{tabular}} & \multicolumn{1}{c}{\begin{tabular}[c]{@{}c@{}}Joint Obj\\ State Acc\end{tabular}} & \multicolumn{1}{c}{\begin{tabular}[c]{@{}c@{}}Joint State \\ IoU@0.5\end{tabular}} & \multicolumn{1}{c}{\begin{tabular}[c]{@{}c@{}}Joint State \\ IoU@0.7\end{tabular}} \\\hline
VDTN                 & 28.0\%                                                                            & 15.3\%                                                                             & 13.1\%                                                                             & 4.0\%                                                                             & 2.2\%                                                                              & 2.0\%                                                                              \\
RNN(V)           & 1.0\%                                                                             & 0.1\%                                                                              & 0.1\%                                                                              & 1.5\%                                                                             & 0.4\%                                                                              & 0.4\%                                                                              \\
RNN(V+D) & 6.8\%                                                                             & 2.6\%                                                                              & 2.3\%                                                                              & 3.7\%                                                                             & 1.8\%                                                                              & 1.6\%                                                                             \\ \hline
\end{tabular}
\caption{Results with and without object representations}
\label{tab:object_results}
\end{table*}

\paragraph{Impacts of dialogue context encoder.}
In Table \ref{tab:dialogue_results}, we observed the benefits of using the Markov process to decode dialogue states based on the dialogue states of the last turn $\mathcal{B}_{t-1}$. 
This strategy allow us to discard parts of dialogue history that is already represented by the state. 
We noted that the optimal design is to use at least $1$ last dialogue turn as the dialogue history. 
In a hypothetical scenario, we applied the oracle $\mathcal{B}_{t-1}$ during test time, and noted the performance is improved significantly. 
This observation indicates the sensitivity of VDTN to a turn-wise auto-regressive decoding process. 

\paragraph{Impacts of frame-level and segment-level sampling.}
As expected, Table \ref{tab:video_results} displays higher performance with higher object limits $N_{obj}$, which increases the chance of detecting the right visual objects in videos. 
We noted performance gains when sampling strides increase up to 24 frames. 
However, in the extreme case, when sampling stride is 300 frames, the performance on temporal slots reduce (as shown by ``Joint State IoU@$p$'').
This raises the issue to sample data more efficiently by balancing between temporal sparsity in videos and state prediction performance. 
We also observed that in a hypothetical scenario with a perfect object perception model, the performance improves significantly, especially on the predictions of discrete slots, although less effect on temporal slots. 

\paragraph{Impacts of object-level representation.}
Table \ref{tab:object_results} reported the results when only segment-level features are used. 
We observed that both VDTN and \emph{RNN(V+D)} are affected significantly, specifically by $24$\% and $3.1$\% ``Joint Obj State Acc'' score respectively. 
Interestingly, we noted that \emph{RNN(V)}, using only video inputs, are not affected by the removal of object-level features. 
These observations indicate that current MM-DST requires object-level information.
We expected that existing 3DCNN models such as ResNeXt still fail to capture such level of granularity. 

\begin{figure*}[htbp]
	\centering
	\resizebox{1.0\textwidth}{!} {
	\includegraphics{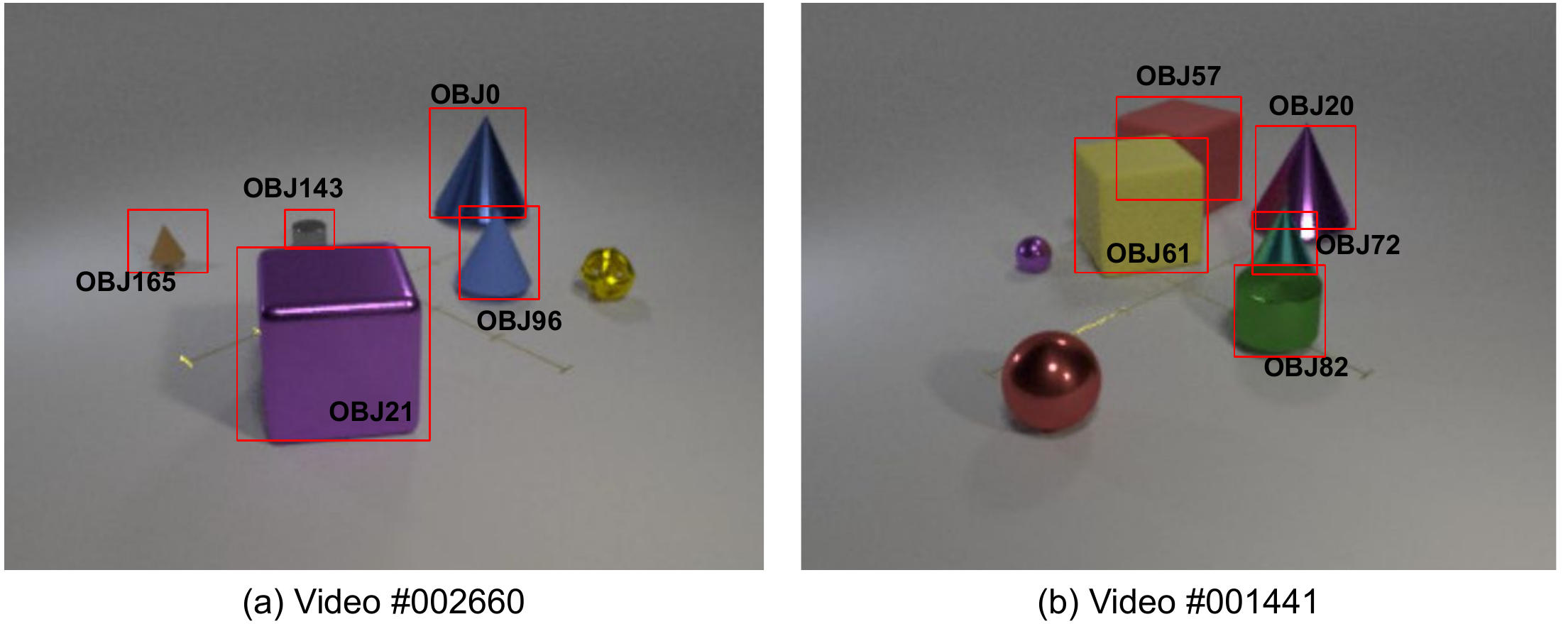}
	}
	\caption{
	Example screenshots of CATER videos for dialogues in Table \ref{tab:example_002660} (Video \#002660) and \ref{tab:example_001441} (Video \#001441). We showed example bounding boxes and their object classes in each video. 
	}
	\label{fig:002660}
\end{figure*}

\paragraph{Qualitative analysis.}
Table \ref{tab:example_002660} and \ref{tab:example_001441} display 2 sample dialogues and state predictions.
We displayed the corresponding video screenshots for these dialogues in Figure \ref{fig:002660}.
To cross-reference between videos and dialogues, we displayed the bounding boxes and their object classes in video screenshots.
These object classes are indicated in ground-truth and decoded dialogue states in dialogues. 
Overall, we noted that VDTN generated temporal slots of start and end time such that the resulting periods better match the ground-truth temporal segments.
VDTN also showed to maintain the dialogue states better from turn to turn. 

\section{Further Discussion}
\label{app:discuss}

\paragraph{Synthetic datasets result in overestimation of real performance and don't translate to real-world usability.}
We agree that the current state accuracy seems to be quite low at about 28\%. However, we want to highlight that state accuracy used in this paper is a very strict metric, which only considers a prediction as correct if it completely matches the ground truth. In DVD, assuming the average 10 objects per video with the set of attributes as in Figure \ref{fig:cater} (+ ‘none’ value in each slot), we can roughly equate the multimodal DST as a 7200-class classification task, each class is a distinct set of objects, each with all possible attribute combinations.  Combined with the cascading error from object perception models, we think the current reported results are reasonable.

Moreover,  we want to highlight that the reported performance of baselines reasonably matches their own capacities in unimodal DST. We can consider Object State F1 as the performance on single-object state and it can closely correlate with the joint state accuracy in unimodal DST (remember that unimodal DST such as MultiWOZ \citep{budzianowski-etal-2018-multiwoz} is only limited to a single object/entity per dialogue). As seen in Table \ref{tab:baseline_results}, the Object State F1 results of TRADE \citep{wu-etal-2019-transferable}, UniConv \citep{le-etal-2020-uniconv}, and NADST \citep{Le2020Non-Autoregressive} are between 46-50\%.
This performance range is indeed not very far off from the performance of these baseline models in unimodal DST in the MultiWOZ benchmark \citep{budzianowski-etal-2018-multiwoz}. 

Finally, we also want to highlight that like other synthetic benchmarks such as CLEVR \citep{johnson2017clevr}, we want to use DVD in this work as a test bed to study and design better multimodal dialogue systems. 
However, we do not intend to use it as a training data for practical systems. 
The DVD-DST benchmark should be used to supplement real-world video-grounded dialogue datasets.

\paragraph{MM-DST in practical applications e.g. with videos of humans.}
While we introduced MM-DST task and VDTN as a new baseline, we noted that the existing results are limited to the synthetic benchmark. 
For instance, in the real world, there would be many identical objects with the same (size, color, material, shape) tuples, which would make the current formulation of dialogue states difficult. 
In such object-driven conversations, we would recommend a dialogue agent not focus on all possible objects in each video frame, but only on a ``focus group'' of objects. 
These objects, required to be semantically different, are topical subjects of the conversations. 

Say we want to scale to a new domain e.g. videos of humans, the first challenge from the current study is the recognition of human objects, which often have higher visual complexity than moving objects as in DVD. We also noted that it is impossible to define all human object classes as in CATER object classes, each of which is unique by its own appearance. To overcome this limitation, we would want to explore multimodal DST with the research of human object tracking, e.g. \citep{fernando2018tracking}, and consider human object identities uniquely defined per video. Another limitation is the definition of slots to track in each human object. While this requires careful considerations, for both practical and ethical reasons, we noted several potential papers that investigate human attributes in dialogues such as human emotions \citep{wang2021contextual}. 
Along these lines, we are excited to see interesting adaptations of multimodal dialogue states grounded on videos of humans. 

\begin{table*}[htbp]
\centering
\tiny
\resizebox{0.95\textwidth}{!} {
\begin{tabular}{llp{10cm}}
\hline
\multirow{5}{*}{\#1}  & HUMAN:         & after the cube 's second rotation , how many other things perform the same sequence of activities as the brown thing ?                                                                                                                                     \\
\cline{2-3}
                         & Gold:          & STAR=102, END=138, (OBJ21, SHAPE, cube), (OBJ165, COLOR, brown)                                                                                                                                                                                            \\
                         & RNN+Attn: & STAR=73, END=73, (OBJ165, COLOR, brown)                                                                                                                                                                                                                    \\
                         & UniConv:       & STAR=73, END=145, (OBJ142, SHAPE, cube), (OBJ165, COLOR, brown)                                                                                                                                                                                            \\
                         & VDTN:          & STAR=97, END=145, (OBJ21, SHAPE, cube), (OBJ165, COLOR, brown)                                                                                                                                                                                             \\\hline
\multirow{6}{*}{\#2}  & SYSTEM:        & 3                                                                                                                                                                                                                                                          \\
                         & HUMAN:         & what about up until now ?                                                                                                                                                                                                                                  \\\cline{2-3}
                         & Gold:          & STAR=102, END=300, (OBJ21, SHAPE, cube), (OBJ165, COLOR, brown)                                                                                                                                                                                            \\
                         & RNN+Attn: & STAR=1, END=300, (OBJ165, COLOR, brown)                                                                                                                                                                                                                    \\
                         & UniConv:       & STAR=145, END=300, (OBJ142, SHAPE, cube), (OBJ165, COLOR, brown)                                                                                                                                                                                           \\
                         & VDTN:          & STAR=97, END=300, (OBJ21, SHAPE, cube), (OBJ165, COLOR, brown)                                                                                                                                                                                             \\\hline
\multirow{6}{*}{\#3}  & SYSTEM:        & 0                                                                                                                                                                                                                                                          \\
                         & HUMAN:         & throughout the whole video, do the large cone and the earlier mentioned brown thing perform the same set of activities ?'                                                                                                                                  \\\cline{2-3}
                         & Gold:          & STAR=1, END=300, (OBJ0, SIZE, large), (OBJ0, SHAPE, cone), (OBJ21, SHAPE, cube), (OBJ165, COLOR, brown)                                                                                                                                                    \\
                         & RNN+Attn: & STAR=1, END=300, (OBJ165, COLOR, brown)                                                                                                                                                                                                                    \\
                         & UniConv:       & STAR=1, END=300, (OBJ142, SHAPE, cube), (OBJ165, COLOR, brown) (OBJ165, SHAPE, cone)                                                                                                                                                                       \\
                         & VDTN:          & STAR=1, END=300, (OBJ21, SHAPE, cube), (OBJ165, COLOR, brown)                                                                                                                                                                                              \\\hline
\multirow{6}{*}{\#4}  & SYSTEM:        & FALSE                                                                                                                                                                                                                                                      \\
                         & HUMAN:         & during the earlier mentioned block 's last rotation , what is the earlier mentioned brown thing doing ?                                                                                                                                                    \\\cline{2-3}
                         & Gold:          & STAR=175, END=200, (OBJ0, SIZE, large), (OBJ0, SHAPE, cone), (OBJ21, SHAPE, cube), (OBJ165, COLOR, brown)                                                                                                                                                  \\
                         & RNN+Attn: & STAR=1, END=300, (OBJ165, COLOR, brown)                                                                                                                                                                                                                    \\
                         & UniConv:       & STAR=265, END=300, (OBJ142, SHAPE, cube), (OBJ165, COLOR, brown) (OBJ165, SHAPE, cone)                                                                                                                                                                     \\
                         & VDTN:          & STAR=241, END=265, (OBJ21, SHAPE, cube), (OBJ165, COLOR, brown)                                                                                                                                                                                            \\\hline
\multirow{6}{*}{\#5}  & SYSTEM:        & no action                                                                                                                                                                                                                                                  \\
                         & HUMAN:         & considering that object, after its slide is there a flying small yellow thing ?'                                                                                                                                                                           \\\cline{2-3}
                         & Gold:          & STAR=291, END=300, (OBJ0, SIZE, large), (OBJ0, SHAPE, cone), (OBJ21, SHAPE, cube), (OBJ165, COLOR, brown)                                                                                                                                                  \\
                         & RNN+Attn: & STAR=1, END=300, (OBJ165, COLOR, brown)                                                                                                                                                                                                                    \\
                         & UniConv:       & STAR=277, END=300, (OBJ142, SHAPE, cube), (OBJ165, COLOR, brown) (OBJ165, SHAPE, cone)                                                                                                                                                                     \\
                         & VDTN:          & STAR=288, END=300, (OBJ21, SHAPE, cube), (OBJ165, COLOR, brown)                                                                                                                                                                                            \\\hline
\multirow{6}{*}{\#6}  & SYSTEM:        & FALSE                                                                                                                                                                                                                                                      \\
                         & HUMAN:         & during the purple thing 's second rotation , what number of big metal things are on the right side of the earlier mentioned brown thing ?                                                                                                                  \\\cline{2-3}
                         & Gold:          & STAR=80, END=102, (OBJ0, SIZE, large), (OBJ0, SHAPE, cone), (OBJ21, COLOR, purple), (OBJ21, SHAPE, cube), (OBJ165, COLOR, brown)                                                                                                                           \\
                         & RNN+Attn: & STAR=1, END=300, (OBJ21, COLOR, purple), (OBJ165, COLOR, brown)                                                                                                                                                                                            \\
                         & UniConv:       & STAR=277, END=300, (OBJ142, SHAPE, cube), (OBJ165, COLOR, brown) (OBJ165, SHAPE, cone)                                                                                                                                                                     \\
                         & VDTN:          & STAR=73, END=97, (OBJ21, COLOR, purple), (OBJ21, SHAPE, cube), (OBJ165, COLOR, brown)                                                                                                                                                                      \\\hline
\multirow{6}{*}{\#7}  & SYSTEM:        & 1                                                                                                                                                                                                                                                          \\
                         & HUMAN:         & what about left of it ?'                                                                                                                                                                                                                                   \\\cline{2-3}
                         & Gold:          & STAR=80, END=102, (OBJ0, SIZE, large), (OBJ0, SHAPE, cone), (OBJ21, COLOR, purple), (OBJ21, MATERIAL, metal), (OBJ21, SHAPE, cube), (OBJ165, COLOR, brown)                                                                                                 \\
                         & RNN+Attn: & STAR=1, END=300, (OBJ21, COLOR, purple), (OBJ165, COLOR, brown)                                                                                                                                                                                            \\
                         & UniConv:       & STAR=73, END=97, (OBJ21, COLOR, purple), (OBJ142, SHAPE, cube), (OBJ165, COLOR, brown) (OBJ165, SHAPE, cone)                                                                                                                                               \\
                         & VDTN:          & STAR=73, END=97, (OBJ21, SIZE, large), (OBJ21, COLOR, purple), (OBJ21, SHAPE, cube), (OBJ143, MATERIAL, metal), (OBJ165, COLOR, brown)                                                                                                                     \\\hline
\multirow{6}{*}{\#8}  & SYSTEM:        & 0                                                                                                                                                                                                                                                          \\
                         & HUMAN:         & before this time window, how many other objects undertake the same types of actions as the earlier mentioned cone ?'                                                                                                                                       \\\cline{2-3}
                         & Gold:          & STAR=1, END=80, (OBJ0, SIZE, large), (OBJ0, SHAPE, cone), (OBJ21, COLOR, purple), (OBJ21, MATERIAL, metal), (OBJ21, SHAPE, cube), (OBJ165, COLOR, brown)                                                                                                   \\
                         & RNN+Attn: & STAR=1, END=300, (OBJ21, COLOR, purple), (OBJ165, COLOR, brown)                                                                                                                                                                                            \\
                         & UniConv:       & STAR=1, END=73, (OBJ21, COLOR, purple), (OBJ21, SHAPE, cube), (OBJ165, COLOR, brown) (OBJ165, SHAPE, cone)                                                                                                                                                 \\
                         & VDTN:          & STAR=1, END=73, (OBJ21, SIZE, large), (OBJ21, COLOR, purple), (OBJ21, SHAPE, cube), (OBJ143, MATERIAL, metal), (OBJ165, COLOR, brown)                                                                                                                      \\\hline
\multirow{6}{*}{\#9}  & SYSTEM:        & 2                                                                                                                                                                                                                                                          \\
                         & HUMAN:         & among them, there is a cone . throughout the whole video, is there any other thing that has the same types of actions as it ?'                                                                                                                             \\\cline{2-3}
                         & Gold:          & STAR=1, END=300, (OBJ0, SIZE, large), (OBJ0, SHAPE, cone), (OBJ21, COLOR, purple), (OBJ21, MATERIAL, metal), (OBJ21, SHAPE, cube), (OBJ96, SHAPE, cone), (OBJ165, COLOR, brown)                                                                            \\
                         & RNN+Attn: & STAR=1, END=300, (OBJ21, COLOR, purple), (OBJ165, COLOR, brown)                                                                                                                                                                                            \\
                         & UniConv:       & STAR=1, END=300, (OBJ21, COLOR, purple), (OBJ21, SHAPE, cube), (OBJ96, COLOR, blue), (OBJ165, COLOR, brown) (OBJ165, SHAPE, cone)                                                                                                                          \\
                         & VDTN:          & STAR=1, END=300, (OBJ21, SIZE, large), (OBJ21, COLOR, purple), (OBJ21, SHAPE, cube), (OBJ143, MATERIAL, metal), (OBJ165, COLOR, brown)                                                                                                                     \\\hline
\multirow{6}{*}{\#10} & SYSTEM:        & FALSE                                                                                                                                                                                                                                                      \\
                         & HUMAN:         & until the end of the blue shiny thing 's last flight , does the earlier mentioned brown object fly as frequently as the cylinder rotates ?                                                                                                                 \\\cline{2-3}
                         & Gold:          & STAR=1, END=228, (OBJ0, SIZE, large),  (OBJ0, COLOR, blue),  (OBJ0, MATERIAL, metal), (OBJ0, SHAPE, cone), (OBJ21, COLOR, purple), (OBJ21, MATERIAL, metal), (OBJ21, SHAPE, cube), (OBJ96, SHAPE, cone), (OBJ143, SHAPE, cylinder), (OBJ165, COLOR, brown) \\
                         & RNN+Attn: & STAR=1, END=300, (OBJ21, COLOR, purple), (OBJ96, COLOR, blue), (OBJ143, SHAPE, cylinder), (OBJ165, COLOR, brown)                                                                                                                                           \\
                         & UniConv:       & STAR=1, END=300, (OBJ21, COLOR, purple), (OBJ96, COLOR, blue), (OBJ143, SHAPE, cylinder), (OBJ165, COLOR, brown)                                                                                                                                           \\
                         & VDTN:          & STAR=1, END=241, (OBJ21, SIZE, large), (OBJ21, COLOR, purple), (OBJ21, SHAPE, cube), (OBJ143, MATERIAL, metal), (OBJ165, COLOR, brown)                                                    \\ \hline                                                                
\end{tabular}
}
\caption{Example predictions from a sample dialogue (CATER video ID \#002660)}
\label{tab:example_002660}
\end{table*}

\begin{table*}[htbp]
\centering
\tiny
\resizebox{0,95\textwidth}{!} {
\begin{tabular}{llp{10cm}}
\hline
\multirow{5}{*}{\#1}  & HUMAN:         & during the red block 's rotation, how many matte things are to the left of the green object ?                                                                                         \\
\cline{2-3}
                         & Gold:          & STAR=143, END=163, (OBJ57, COLOR, red), (OBJ57, SHAPE, cube), (OBJ82, COLOR, green)                                                                                                   \\
                         & RNN+Attn: & STAR=37, END=109, (OBJ57, SHAPE, cube), (OBJ82, COLOR, green), (OBJ91, COLOR, red)                                                                                                    \\
                         & UniConv:       & STAR=73, END=109, (OBJ57, MATERIAL, rubber), (OBJ82, COLOR, green), (OBJ91, COLOR, red)                                                                                               \\
                         & VDTN:          & STAR=73, END=109, (OBJ57, COLOR, red), (OBJ57, SHAPE, cube), (OBJ82, COLOR, green)                                                                                                    \\\hline
\multirow{6}{*}{\#2}  & SYSTEM:        & 2                                                                                                                                                                                     \\
                         & HUMAN:         & how about in front of it ?                                                                                                                                                            \\\cline{2-3}
                         & Gold:          & STAR=143, END=163, (OBJ57, COLOR, red), (OBJ57, SHAPE, cube), (OBJ82, COLOR, green)                                                                                                   \\
                         & RNN+Attn: & STAR=37, END=25, (OBJ82, COLOR, green), (OBJ91, COLOR, red)                                                                                                                           \\
                         & UniConv:       & STAR=73, END=109, (OBJ57, MATERIAL, rubber), (OBJ82, COLOR, green), (OBJ91, COLOR, red)                                                                                               \\
                         & VDTN:          & STAR=73, END=109, (OBJ57, COLOR, red), (OBJ57, SHAPE, cube), (OBJ82, COLOR, green)                                                                                                    \\\hline
\multirow{6}{*}{\#3}  & SYSTEM:        & 0                                                                                                                                                                                     \\
                         & HUMAN:         & before this time window, how many moving things are there ?                                                                                                                           \\\cline{2-3}
                         & Gold:          & STAR=1, END=143, (OBJ57, COLOR, red), (OBJ57, SHAPE, cube), (OBJ82, COLOR, green)                                                                                                     \\
                         & RNN+Attn: & STAR=1, END=1, (OBJ82, COLOR, green), (OBJ91, COLOR, red)                                                                                                                             \\
                         & UniConv:       & STAR=1, END=73, (OBJ57, COLOR, red), (OBJ57, MATERIAL, rubber), (OBJ82, COLOR, green), (OBJ91, COLOR, red)                                                                            \\
                         & VDTN:          & STAR=1, END=73, (OBJ57, COLOR, red), (OBJ57, SHAPE, cube), (OBJ82, COLOR, green)                                                                                                      \\\hline
\multirow{6}{*}{\#4}  & SYSTEM:        & 5                                                                                                                                                                                     \\
                         & HUMAN:         & among them, there is a purple thing . during the whole video, how many other things with the same sequence activities performed by it ?                                               \\\cline{2-3}
                         & Gold:          & STAR=1, END=168, (OBJ20, COLOR, purple), (OBJ57, COLOR, red), (OBJ57, SHAPE, cube), (OBJ82, COLOR, green)                                                                             \\
                         & RNN+Attn: & STAR=1, END=169, (OBJ82, COLOR, green), (OBJ91, COLOR, red)                                                                                                                           \\
                         & UniConv:       & STAR=1, END=169, (OBJ57, COLOR, red), (OBJ57, MATERIAL, rubber), (OBJ82, COLOR, green), (OBJ91, COLOR, red), (OBJ152, COLOR, purple)                                                  \\
                         & VDTN:          & STAR=1, END=169, (OBJ20, COLOR, purple), (OBJ57, COLOR, red), (OBJ57, SHAPE, cube), (OBJ82, COLOR, green)                                                                             \\\hline
\multirow{6}{*}{\#5}  & SYSTEM:        & 1                                                                                                                                                                                     \\
                         & HUMAN:         & how about up until now ?                                                                                                                                                              \\\cline{2-3}
                         & Gold:          & STAR=1, END=247, (OBJ20, COLOR, purple), (OBJ57, COLOR, red), (OBJ57, SHAPE, cube), (OBJ82, COLOR, green)                                                                             \\
                         & RNN+Attn: & STAR=1, END=253, (OBJ82, COLOR, green), (OBJ91, COLOR, red)                                                                                                                           \\
                         & UniConv:       & STAR=1, END=253, (OBJ57, COLOR, red), (OBJ57, MATERIAL, rubber), (OBJ82, COLOR, green), (OBJ91, COLOR, red), (OBJ152, COLOR, purple)                                                  \\
                         & VDTN:          & STAR=1, END=253, (OBJ20, COLOR, purple), (OBJ57, COLOR, red), (OBJ57, SHAPE, cube), (OBJ82, COLOR, green)                                                                             \\\hline
\multirow{6}{*}{\#6}  & SYSTEM:        & 0                                                                                                                                                                                     \\
                         & HUMAN:         & during the earlier mentioned purple thing 's flight , is there a flying green object ?                                                                                                \\\cline{2-3}
                         & Gold:          & STAR=45, END=69, (OBJ20, COLOR, purple), (OBJ57, COLOR, red), (OBJ57, SHAPE, cube), (OBJ82, COLOR, green)                                                                             \\
                         & RNN+Attn: & STAR=1, END=241, (OBJ82, COLOR, green), (OBJ91, COLOR, red)                                                                                                                           \\
                         & UniConv:       & STAR=73, END=97, (OBJ57, COLOR, red), (OBJ57, MATERIAL, rubber), (OBJ82, COLOR, green), (OBJ91, COLOR, red), (OBJ152, COLOR, purple)                                                  \\
                         & VDTN:          & STAR=73, END=97, (OBJ20, COLOR, purple), (OBJ57, COLOR, red), (OBJ57, SHAPE, cube), (OBJ82, COLOR, green)                                                                             \\\hline
\multirow{6}{*}{\#7}  & SYSTEM:        & FALSE                                                                                                                                                                                 \\
                         & HUMAN:         & during the yellow thing 's second rotation , does the cylinder rotate more frequently than the aforementioned purple object flies ?                                                   \\\cline{2-3}
                         & Gold:          & STAR=74, END=98, (OBJ20, COLOR, purple), (OBJ57, COLOR, red), (OBJ57, SHAPE, cube), (OBJ61, COLOR, yellow), (OBJ82, COLOR, green), (OBJ82, SHAPE, cylinder)                           \\
                         & RNN+Attn: & STAR=1, END=241, (OBJ61, COLOR, yellow), (OBJ82, COLOR, green), (OBJ82, SHAPE, cylinder), (OBJ91, COLOR, red)                                                                         \\
                         & UniConv:       & STAR=73, END=97, (OBJ57, COLOR, red), (OBJ57, MATERIAL, rubber), (OBJ82, COLOR, green), (OBJ91, COLOR, red), (OBJ152, COLOR, purple)                                                  \\
                         & VDTN:          & STAR=73, END=97, (OBJ20, COLOR, purple), (OBJ57, COLOR, red), (OBJ57, SHAPE, cube), (OBJ61, COLOR, yellow), (OBJ82, COLOR, green)                                                     \\\hline
\multirow{6}{*}{\#8}  & SYSTEM:        & TRUE                                                                                                                                                                                  \\
                         & HUMAN:         & during the aforementioned cube 's rotation , what types of actions does the cone left of the aforementioned yellow object undertake ?                                                 \\\cline{2-3}
                         & Gold:          & STAR=143, END=163, (OBJ20, COLOR, purple), (OBJ57, COLOR, red), (OBJ57, SHAPE, cube), (OBJ61, COLOR, yellow), (OBJ72, SHAPE, cone), (OBJ82, COLOR, green), (OBJ82, SHAPE, cylinder)   \\
                         & RNN+Attn: & STAR=1, END=193, (OBJ20, COLOR, purple), (OBJ20, SHAPE, cone), (OBJ57, COLOR, red), (OBJ61, COLOR, yellow), (OBJ82, COLOR, green), (OBJ82, SHAPE, cylinder), (OBJ91, COLOR, red)      \\
                         & UniConv:       & STAR=73, END=97, (OBJ57, MATERIAL, rubber), (OBJ72, SHAPE, cone), (OBJ82, COLOR, green), (OBJ82, SHAPE, cylinder), (OBJ91, COLOR, red), (OBJ152, COLOR, purple)                       \\
                         & VDTN:          & STAR=73, END=97, (OBJ20, COLOR, purple), (OBJ20, SHAPE, cone), (OBJ57, COLOR, red), (OBJ57, SHAPE, cube), (OBJ61, COLOR, yellow), (OBJ82, COLOR, green)                               \\\hline
\multirow{6}{*}{\#9}  & SYSTEM:        & flying                                                                                                                                                                                \\
                         & HUMAN:         & throughout the whole video, is there anything else that performs the same set of activities as the earlier mentioned yellow thing ?                                                   \\\cline{2-3}
                         & Gold:          & STAR=1, END=247, (OBJ20, COLOR, purple), (OBJ57, COLOR, red), (OBJ57, SHAPE, cube), (OBJ61, COLOR, yellow), (OBJ72, SHAPE, cone), (OBJ82, COLOR, green), (OBJ82, SHAPE, cylinder)     \\
                         & RNN+Attn: & STAR=1, END=241, (OBJ20, COLOR, purple), (OBJ20, SHAPE, cone), (OBJ57, COLOR, red), (OBJ61, COLOR, yellow), (OBJ82, COLOR, green), (OBJ82, SHAPE, cylinder), (OBJ91, COLOR, red)      \\
                         & UniConv:       & STAR=1, END=253, (OBJ57, MATERIAL, rubber), (OBJ57, SHAPE, cube), (OBJ72, SHAPE, cone), (OBJ82, COLOR, green), (OBJ82, SHAPE, cylinder), (OBJ91, COLOR, red), (OBJ152, COLOR, purple) \\
                         & VDTN:          & STAR=1, END=253, (OBJ20, COLOR, purple), (OBJ20, SHAPE, cone), (OBJ57, COLOR, red), (OBJ57, SHAPE, cube), (OBJ61, COLOR, yellow), (OBJ82, COLOR, green)                               \\\hline
\end{tabular}
}
\caption{Example predictions from a sample dialogue (CATER video ID \#001441)}
\label{tab:example_001441}
\end{table*}

\end{document}